\documentclass{article} %

\usepackage{amsmath,amsfonts,bm}

\def\eqref#1{equation~\ref{#1}}

\def\1{\bm{1}}

\DeclareMathAlphabet{\mathsfit}{\encodingdefault}{\sfdefault}{m}{sl}
\SetMathAlphabet{\mathsfit}{bold}{\encodingdefault}{\sfdefault}{bx}{n}

\usepackage{hyperref}
\usepackage{url}
\usepackage{cleveref} 
\usepackage{graphicx}
\usepackage{xcolor} 
\usepackage{enumitem} 
\usepackage{booktabs}
\usepackage{adjustbox}
\usepackage{wrapfig}
\usepackage{subcaption}
\usepackage{float}
\usepackage{fullpage}
\usepackage{parskip}
\usepackage{natbib}
\usepackage[table]{xcolor}

\usepackage[framemethod=TikZ]{mdframed}
\mdfdefinestyle{MyFrame}{%
    linecolor=gray!50,
    linewidth=1pt,
    backgroundcolor=gray!5,
    roundcorner=5pt,
    innertopmargin=\baselineskip,
    innerbottommargin=\baselineskip,
    innerrightmargin=15pt,
    innerleftmargin=15pt,
    skipabove=\baselineskip,
    skipbelow=\baselineskip,
    frametitlefont=\bfseries,
}

\title{CTRL-Rec: Controlling Recommender Systems\\With Natural Language}

\author{
Micah Carroll,$^{1}$ Adeline Foote,$^{1}$ Kevin Feng,$^{2}$ Marcus Williams,$^{1}$ \\
Anca Dragan,$^{3}$ W. Bradley Knox,*$^{, 4}$ Smitha Milli*$^{, 5}$ \\
{\small $^{1}$MATS \quad $^{2}$University of Washington \quad $^{3}$UC Berkeley \quad $^{4}$UT Austin \quad $^{5}$FAIR at Meta }
}
\date{}

\begin{document}

\maketitle

\begin{abstract}
    When users are dissatisfied with recommendations from a recommender system, they often lack fine-grained controls for changing them. Large language models (LLMs) offer a solution by allowing users to guide their recommendations through natural language \textit{requests} (e.g., ``I want to see respectful posts with a different perspective than mine"). We propose a method, \textbf{CTRL-Rec}, that allows for natural language control of traditional recommender systems in real-time with computational efficiency. Specifically, at training time, we use an LLM to simulate whether users would approve of items based on their language requests, and we train embedding models that approximate such simulated judgments. We then integrate these user-request-based predictions into the standard weighting of signals that traditional recommender systems optimize. At deployment time, we require only a single LLM embedding computation per user request, allowing for real-time control of recommendations. In experiments with the MovieLens dataset, our method consistently allows for fine-grained control across a diversity of requests. In a study with 19 Letterboxd users, we find that CTRL-Rec significantly enhances users’ sense of control and satisfaction with their recommendations, without reducing engagement.
\end{abstract}

\section{Introduction}

In this work, we propose \textit{\textbf{CTRL-Rec}}, a method that integrates natural language controls into traditional recommender systems. CTRL-Rec allows for balancing between users' explicitly stated preferences and their engagement signals, and importantly, is capable of directly influencing the retrieval stage of modern recommender systems, rather than being confined to post-retrieval ranking.

\begin{figure}[t]
    \vspace{-1em}
    \centering
    \includegraphics[width=0.8\linewidth]{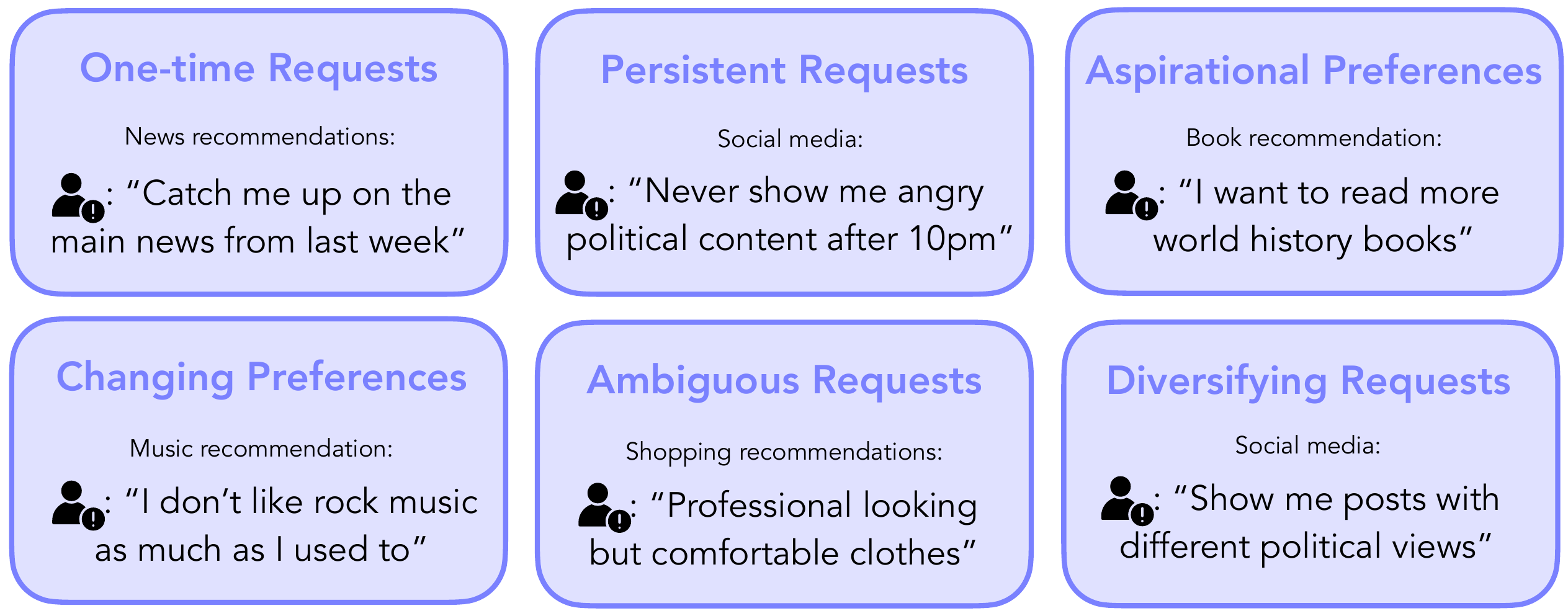}
    \vspace{-0.6em}
    \caption{CTRL-Rec has the potential to be able to handle many kinds of user requests, unlocking many novel forms of user control.}
    \label{fig:request-types} 
    \vspace{-1.4em}
\end{figure}

\textbf{Motivation.} Natural language could serve as an intuitive and flexible interface for controlling recommender systems~\citep{friedman_leveraging_2023,malki2025bonsaiintentionalpersonalizedsocial}. Imagine that a user could simply say ``I want content that helps me learn about other perspectives," (or any of the other requests in \Cref{fig:request-types}) and have their recommendations update immediately in real-time. These controls could act as an important counter-balance to the engagement-focused\footnote{Modern recommender systems are primarily based on optimizing user engagement, e.g., likes, clicks, replies, etc. Platforms do also incorporate non-engagement signals like user controls and surveys to some extent~\citep{cunningham2025ranking}, however, current integrations are typically limited by the sparsity of existing non-engagement signals.} nature of modern recommender systems, giving individuals greater agency to align the recommender system with reflective or aspirational preferences that aren't captured by their engagement history~\citep{ekstrand2016,kleinberg2024inversion,morewedge2023human,lazar_moral_2024,rezk2024,lukoff2021,lukoff2023,fangyi2025,zhang2022,feng2024}. For instance, a user might want to see thoughtful long-form history videos, even if haven't typically engaged with such content. More intentional curation using stated preferences could also allow users to counter divisive but undesired content amplified by engagement-optimizing  algorithms~\citep{milli2025engagementusersatisfactionamplification,rathje2024people}.

Large language models (LLMs) offer promising capabilities for fulfilling such requests~\citep{malki2025bonsaiintentionalpersonalizedsocial,kolluri2026alexandrialibrarypluralisticvalues,jahanbakhsh2025valuealignmentsocialmedia}. However, two key practical challenges remain in fully integrating natural language with conventional recommender systems. \textit{First}, it is essential to be able to balance user engagement with stated preferences, since users will articulate only a small subset of their preferences, requiring the system to infer others from their behavior~\citep{malki2025bonsaiintentionalpersonalizedsocial}. \textit{Second}, while LLMs could be used to directly re-rank a small set of items already present in a user’s feed~\citep{kolluri2026alexandrialibrarypluralisticvalues,jia2024,piccardi2024social,jahanbakhsh2025valuealignmentsocialmedia}, if the initial pool of items is too limited, then re-ranking alone may not sufficiently reflect the user’s expressed preferences. Therefore, it is important to enable natural language controls that can influence the retrieval of items from the outset—a task that is computationally challenging to directly apply LLMs to, given that the retrieval stage often involves billions of items.

\textbf{CTRL-Rec.} Our work tackles both challenges. First, to balance engagement with stated preferences, we overcome the ``type mismatch'' between conversational and traditional recommender systems. While traditional recommenders focus on predicting user-item interactions—such as whether a user will like a specific post—natural language requests are often broad and free-form, potentially applying to many items at once (e.g., “I want to see posts that are funny and witty but not mean”). To bridge this gap, we use LLMs to simulate users' judgments of particular items based upon their natural language requests. These user-request-based predictions can then be integrated into the standard weighting of signals that traditional recommender systems optimize, allowing for a balance of stated preferences and engagement. 

Second, building upon ideas from dense retrieval~\citep{izacard2020,karpukhin_dense_2020,khattab2020}, we enable real-time control by distilling LLM-generated judgments into a dual-encoder (two-tower) architecture for computational efficiency. This approach enables natural language controls to directly influence which items are retrieved, rather than being limited to the final ranking stage. While naively computing the simulated user-item judgments would require $m$ LLM queries for each new user request (where $m$ could be in the billions or trillions), CTRL-Rec instead requires only a single LLM embedding computation per user request.

\begin{figure}[t]
    \centering
    \vspace{-1em}
    \includegraphics[width=\linewidth]{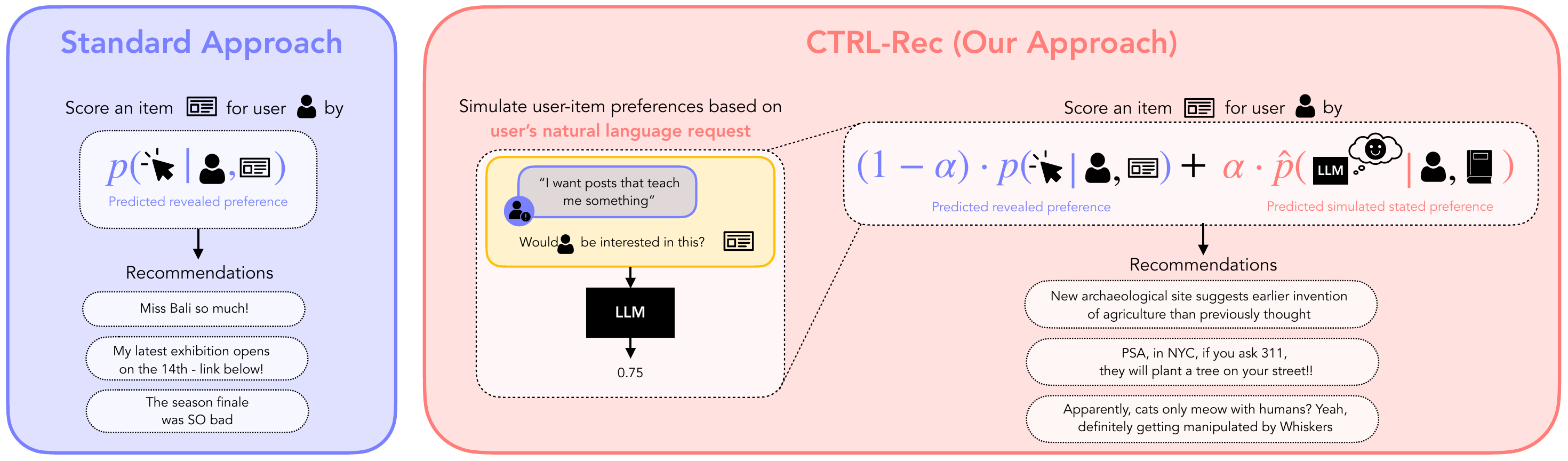}
    \vspace{-1.5em}
    \caption{Our approach differs from standard engagement recommenders by allowing to optimize for both revealed preferences (through engagement signals) and stated preferences (through natural language requests). This allows users to explicitly control their recommendations while maintaining the benefits of engagement-based recommendation.}
    \vspace{-1.4em}
    \label{fig:method} 
\end{figure}

In summary, our contributions are:
\begin{enumerate}[leftmargin=*, itemsep=0pt, topsep=0pt, parsep=4pt, partopsep=0pt]
    \item \textbf{A scalable method for integrating language-based controls} into traditional recommender systems, allowing for real-time control of recommender systems (\Cref{fig:method}). Only one LLM embedding computation is required per user request.
    \item \textbf{Empirical validation on MovieLens} using both genre-specific and subjective, open-ended controls. We find that our approach effectively steers recommendations according to user requests while maintaining engagement quality.
    \item \textbf{Human study with Letterboxd users.} In a study with 19 Letterboxd users, we find that CTRL-Rec was positively received by users and significantly enhanced users’ sense of control and satisfaction with recommendations compared to traditional controls.
\end{enumerate}

\vspace{-0.5em}
\section{Related Work}
\vspace{-0.5em}

\textbf{User Control of Recommender Systems.} Traditional interfaces for allowing users to better align recommender systems with their stated preferences~\citep{milli2021optimizing,kleinberg2024challenge,agarwal2024system,yang2019} include controls such as ``See less often'' buttons or domain-specific controls such as genre toggles. However, research has shown that these controls are rarely used~\citep{cunningham2025ranking}, are often ineffective when they are used~\citep{mozilla_youtube_regrets}, and that users still frequently feel a lack of agency or control over their recommender systems~\citep{lukoff2021,lukoff2023}. Recent work by \citet{kolluri2026alexandrialibrarypluralisticvalues} and \citet{jahanbakhsh2025valuealignmentsocialmedia} aims to broaden user control by introducing value-based toggles, using LLMs to rerank social media content according to selected values (e.g., “cheerful,” “knowledge”, ``tradition''). In contrast, our approach enables users to specify their own free-form values and applies these preferences directly to the retrieval process, rather than only to post-retrieval ranking. Relatedly, \citet{malki2025bonsaiintentionalpersonalizedsocial} leverage LLMs to create custom Bluesky feeds. However, for future work, they emphasize the importance of real-time, in-context control: ``Rather than treating feedbuilding as a separate setup activity, future systems should support lightweight, iterative changes directly within the feed interface.'' CTRL-Rec’s efficiency supports this kind of immediate, interactive control.

\textbf{LLMs for Recommender Systems and Retrieval.} There has been an increasing amount of research focused on leveraging large language models for recommendation \citep{wu_survey_2024,wang_towards_2024,lin_how_2024,huang_foundation_2024} and information retrieval more broadly \citep{karpukhin_dense_2020,khattab2020}. Our work sits somewhere between recommendation and retrieval, as we provide a way for users to provide natural language descriptions of their preferences which are not necessarily one-time searches (e.g. ``never show me angry content about U.S. politics'').

Prior research has directly used LLMs to score the relevance of items to users based on natural language preference strings~\citep{sanner_large_2023}. However, due to the high computational cost associated with LLMs, they are limited to scoring or re-ranking only a small candidate set of items~\citep{jia2024,kolluri2026alexandrialibrarypluralisticvalues}. More recent work in generative retrieval trains models to directly generate the identifier of the next item the user will interact with~\citep{rajput2023,li2025}, however, this approach also suffers from high computational cost and scalability issues to large corpora~\citep{pradeep-etal-2023-generative}. In contrast, our approach enables efficient real-time control by distilling synthetic LLM judgments. The general idea of generating synthetic labels via an LLM and then distilling has been applied before in dense retrieval~\citep{izacard2020,bonifacio_inpars_2022,huang_pairdistill_2024},
although with different types of judgments and for different purposes than our goal here (e.g. scoring ``relevance'' of answers to user questions). Note that, while in this work we distill the LLM judgments into a dual-encoder architecture~\citep{karpukhin_dense_2020,bromley1993}, our general framework is agnostic to the exact distillation method used and could leverage other methods from dense retrieval.

The goal of our work is to give users explicit control over standard recommendation interfaces through natural language requests. In this way, our work contrasts from other research on providing recommendations limited to a chat interface~\citep{gao_chat-rec_2023,friedman_leveraging_2023,he_large_2023} that do not affect traditional recommender systems. It also contrasts from work that uses natural language features or profiles for recommendation~\citep{mysore2023,kim2024reviewdrivenpersonalizedpreferencereasoning,paischer_preference_2024}, but where these profiles are extracted or learned, rather than provided by the user as a means of control. Extracted natural language profiles are also common in more recent works on scrutable recommender systems \citep{radlinski_natural_2022, penaloza_tears_2025, ramos_transparent_2024, gao_end--end_2025,mysore_editable_2023}, which also allow for user control via editing of said profiles. However, CTRL-Rec differs both in methodology for connecting natural language to recommendations, and extends the paradigm to accommodate immediate requests (which are non-persistent), enabling more dynamic control. %

\textbf{}

\begin{figure}[t]
    \vspace{-1.5em}
    \centering
    \includegraphics[width=1\linewidth]{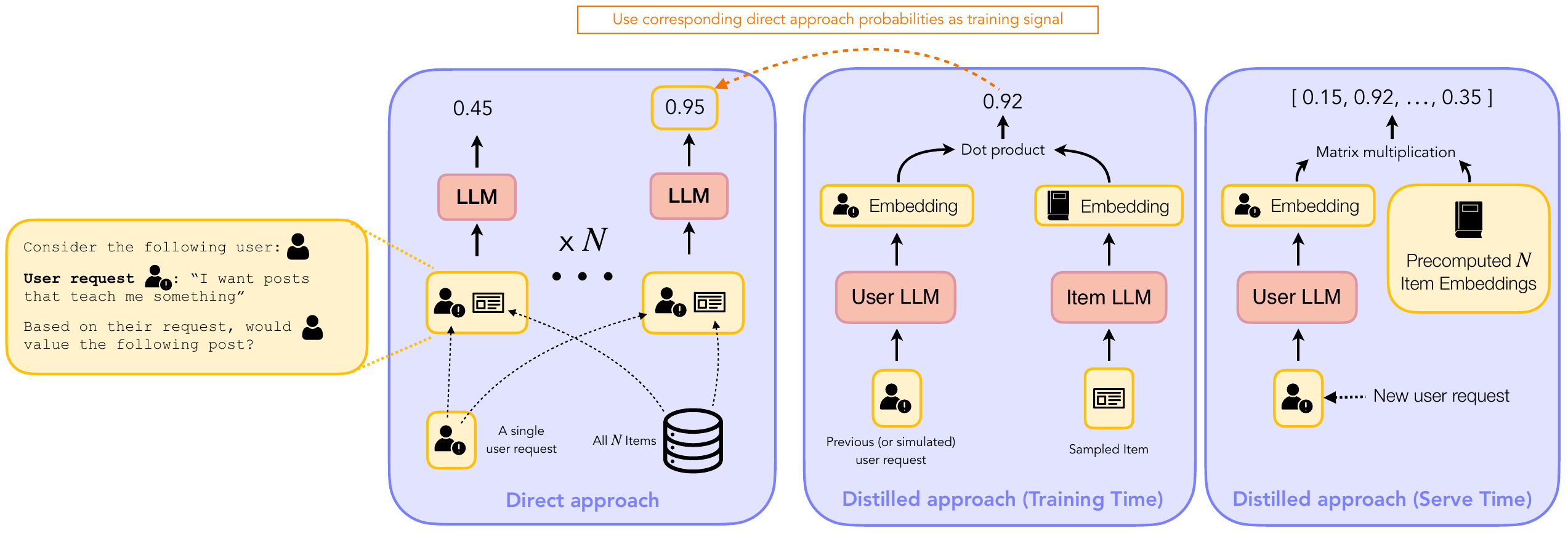}
    \vspace{-1em}
    \caption{\textbf{Overview of our method.} We train a distilled model to approximate LLM judgements of whether a user would like a specific item based on their natural language request. This obviates the need for LLM calls (apart from a single request embedding operation) at test time.}
    \label{fig:model} 
    \vspace{-1em}
\end{figure}

\section{CTRL-Rec: Control Through Language for Recommendations}

\vspace{-0.5em}

Our system integrates natural language control into traditional recommender systems by introducing a novel preference prediction component that estimates how well items align with user-specified requests.

\textbf{Traditional Recommender Scoring.}
Modern recommender systems typically optimize a weighted linear combination of different user-item signals~\citep{milli2023choosing,cunningham2025ranking,smith_2021,Twitter_2023}.
Generally, for a given user $u$, the score of an item $i$ is computed as
\begin{equation}\label{eq:base_score}
    \text{score}_{\text{base}}(u,i) = \sum_{k} w_k f_k(u,i)
\end{equation}
where each function $f_k$ returns the value of the $k$-th user-item signal and $w_k$ is the weight on that signal. The different signals $f_k(u, i)$ are typically probabilities or scores for how likely the user $u$ is to interact with item $i$ in different ways (e.g., liking, commenting). Some signals may also be independent of the user $u$, for example, a signal for the likelihood that the item $i$ violates content moderation standards.

\textbf{Request-aware Recommender Scoring in CTRL-Rec.}
Our key insight is that we can naturally incorporate natural language controls into this framework by explicitly incorporating the user's natural language request $r$ as an independent signal. Our updated scoring function is defined as:
\begin{equation}\label{eq:realm_score}
    \text{score}_{\text{CTRL-Rec}}(u,i,\textcolor{orange}{r}) = \underbrace{\text{score}_{\text{base}}(u,i)}_{\text{revealed preference + other signals}} \,\,+ \,\,w_{\text{control}} \underbrace{\textcolor{orange}{v(u,i,r)}}_{\text{stated preference}}
\end{equation}
Here, the function $v(u,i,r)$ represents \textit{how satisfied user $u$ would be with item $i$ given their request $r$}, and the parameter $w_{\text{control}}$ modulates the trade-off between engagement (the user's revealed preferences) and the user's preferences stated in natural language. The nature of the user request $r$ can be very flexible. It could be a simple immediate request or search, or represent more complex, long-term preferences that users would like the recommender system to \textit{always} follow (e.g. ``Never recommend me war movies'') or only follow \textit{under certain conditions} (e.g. ``I do not want to see political content after 8pm'').\footnote{For the latter case, the user state $u$ must contain sufficient information.}

Platforms typically select the weights on different signals through A/B tests~\citep{cunningham2025ranking,Twitter_2023,milli2023choosing}. By tuning the $w_{\text{control}}$ weight, the platform (or potentially even the user)\footnote{Platforms could easily expose an interface that allows the user to select how to trade-off between their language request (stated preferences) and engagement (revealed preferences).} can find an effective balance between the user's revealed and stated preferences. Also, while \Cref{eq:realm_score} uses a linear interpolation between engagement and stated preferences to align with industry practice, other promising combination strategies are possible~\citep{milli2021optimizing}, such as threshold-based filtering or multiplicative scoring. Empirically, we found that directly interpolating $\text{score}_\text{base}$ and $v(u, i, r)$ performed worse than first ranking all movies by each metric separately and then interpolating between those ranks as above -- so we follow this approach in our experiments. We suspect that this was due to two functions having very differently tailed distributions within [0, 1], which ranks are more robust to.

\subsection{Value Prediction}\label{subsec:value-pred}
The key challenge for our method is estimating $v(u,i,r)$---how well an item aligns with a user's request---and doing so efficiently. A naive approach, which we call \textbf{direct LLM scoring}, is to ask an LLM directly to rate how well each candidate item matches the user's request. 
However, the direct LLM scoring is computationally infeasible for at-scale deployments: it requires performing $m$ LLM queries per user request, where $m$ is the number of items in the candidate pool.\footnote{One might wonder if it is possible to simply do one LLM query per user request but ask the LLM to generate ratings for all $m$ items. When $m$ is large, this request requires a long context, particularly if the items are also represented by richer natural language descriptions as in \Cref{sec:exps}. The long context makes this approach either infeasible or slow.}

To address this, we can distill the LLM's judgments into a more efficient scoring function---what we call the \textbf{distilled approach}---that reduces the computation needed at deployment time from $m$ LLM queries per user request to just one LLM embedding computation. Specifically, we fine-tune LLM embedding models $f$ and $g$ to approximate LLM judgments: $v(u,i,r) \approx f(u,r)^T g(i)$. The item embeddings $g(i)$ can be computed offline, meaning that at deployment time, only one LLM embedding computation per user request (plus a matrix multiplication) is needed. Moreover, as we show in \Cref{subsec:efficiency} this approach can be further optimized by batching simultaneous user requests together. While the idea of using LLM embeddings for engagement prediction has been explored before, we are not aware of prior work on distilling LLMs for this specific item-value prediction task which is conditioned on user's natural language requests.

\section{Experiments} \label{sec:exps}

To evaluate the benefits of CTRL-Rec—which enables both (a) real-time control at the retrieval stage and (b) balancing engagement with stated preferences—we require access to a large and up-to-date inventory of items, along with user engagement data. Although CTRL-Rec’s advantages are likely to be most evident in high-volume, rapidly changing content domains such as social media, news recommendation, or e-commerce, these domains typically lack publicly available, timely large-scale item inventories with sufficient engagement data. Access to a recent inventory is particularly important, as we aim to test our method in user studies, and users would expect recommendations to be relevant and current.\footnote{For example, while the MIND dataset~\citep{wu-etal-2020-mind} is a publicly available news dataset with impression logs, it does not contain up-to-date news data suitable for a user study.}

In light of this, we test CTRL-Rec in the movie recommendation domain, which has standard large-scale datasets of items and user interactions, including recent movies that remain relevant to users today. We evaluate CTRL-Rec's performance through two types of complementary experiments\footnote{Meta served in only an advisory capacity. The research participant study and all experiments were led and implemented by UC Berkeley, UW, and UT Austin.}: (a) a reachability experiment~\citep{dean_recommendations_2020,curmei21}, and (b) a human study involving users of Letterboxd, a movie ratings platform. In the reachability experiment, we randomly select two users from the MovieLens dataset—a source user and a target user—and assess how easily the source user's recommendations can be adjusted to resemble those of the target user. This large-scale experiment evaluates CTRL-Rec's potential steering to \emph{any} observed preferences in the dataset. By contrast, our human study is smaller in scale but focuses on whether CTRL-Rec help real users steer their recommendations with the \emph{specific} preferences they have in mind. When referring to engagement-based recommendations, across all of our experiments we use the SAR (Sequential Association Rules) algorithm, as implemented by Microsoft's \texttt{recommender} module~\citep{graham2019}.

\Cref{fig:qual-examples} shows qualitative examples returned by CTRL-Rec for three different queries. Despite the nuanced and open-ended nature of some of these requests, CTRL-Rec effectively produces relevant results. During deployment, each query requires only one LLM embedding computation, allowing for real-time control of recommendations.

\subsection{Training Details} \label{sec:exp-details}

CTRL-Rec was trained on 220,000 simulated user preferences generated from the MovieLens 32M dataset, which contains $\sim$32m ratings by $\sim$330k unique users across $\sim$87k movies. For each movie in the dataset, we prompted Claude 3.5 to generate a summary as described in \Cref{appendix:prompts}, which is used as input to the CTRL-Rec as a richer representation of each movie than just the title. In other domains, one could use available metadata one has for the item. Our training data generation process consisted of two main stages: (1) request generation, and (2) preference scoring.

\textbf{Request Generation.} We first generated a diverse set of natural language requests by sampling users from the MovieLens dataset and instructing gemini-2.5-flash-preview-05-20 to generate requests across 10 different categories designed to span the range of ways users might express movie preferences. These categories include one-time situational requests, long-term persistent preferences, aspirational requests for personal growth, similarity-based comparisons, and various types of filtering requests ranging from simple to complex logical operations. This approach aims at broad coverage of request types that users might realistically make, though in practice a deployed system could be trained on the actual distribution of user requests.

\textbf{Preference Scoring.} For each simulated preference, we randomly sample movies from the dataset and used Llama 3.1 70B to score how well each movie matched the specific preference request. These LLM-generated scores serve as the target preferences for the distillation step described in our method (see \Cref{appendix:llm_scoring} for details on extracting scores from LLMs). This process was repeated across many user-preference-movie combinations to generate our final training dataset of 220,000 preference examples. See \Cref{app:training-categories} for detailed descriptions of the 10 request categories and the prompts used to generate each type.

\begin{figure}[t!]
\vspace{-2em}
\centering
\caption{Qualitative examples of CTRL-Rec recommendations for three diverse natural language requests ($w_\text{control} = 0.995$ \& no engagement history).}
\vspace{-0.5em}

\begin{minipage}[t]{0.325\textwidth}
\begin{mdframed}[
    linewidth=1pt,
    linecolor=blue!60,
    backgroundcolor=blue!8,
    roundcorner=3pt,
    innertopmargin=2pt,
    innerbottommargin=2pt,
    innerrightmargin=2pt,
    innerleftmargin=2pt,
    skipabove=0pt,
    skipbelow=0pt
]
\footnotesize
\vbox to 1.2cm{
\centering \textbf{Request}\\
\centering \textit{``Movies with inclement weather''}
\vfill}
\end{mdframed}
\begin{mdframed}[
    linewidth=1pt,
    linecolor=gray!50,
    backgroundcolor=gray!5,
    roundcorner=3pt,
    innertopmargin=2pt,
    innerbottommargin=2pt,
    innerrightmargin=2pt,
    innerleftmargin=2pt,
    skipabove=0pt,
    skipbelow=0pt
]
\footnotesize
\vbox to 2.4cm{
\centering \textbf{CTRL-Rec Feed}\\
\raggedright Survive! (1976)\\
The Day After Tomorrow (2004)\\
The Perfect Storm (2000)\\
Deep Impact (1998)\\
The Ghost and the Darkness (1996)
\vfill}
\end{mdframed}
\end{minipage}
\hfill
\begin{minipage}[t]{0.325\textwidth}
\begin{mdframed}[
    linewidth=1pt,
    linecolor=blue!60,
    backgroundcolor=blue!8,
    roundcorner=3pt,
    innertopmargin=2pt,
    innerbottommargin=2pt,
    innerrightmargin=2pt,
    innerleftmargin=2pt,
    skipabove=0pt,
    skipbelow=0pt,
]
\footnotesize
\vbox to 1.2cm{
\centering \textbf{Request}\\
\centering \textit{``I'd like to see more movies with the vibe as Her but not about AI''}
\vfill}
\end{mdframed}
\begin{mdframed}[
    linewidth=1pt,
    linecolor=gray!50,
    backgroundcolor=gray!5,
    roundcorner=3pt,
    innertopmargin=2pt,
    innerbottommargin=2pt,
    innerrightmargin=2pt,
    innerleftmargin=2pt,
    skipabove=0pt,
    skipbelow=0pt,
]
\footnotesize
\vbox to 2.4cm{
\centering \textbf{CTRL-Rec Feed}\\
\raggedright Garden State (2004)\\
Once (2006)\\
Lars and the Real Girl (2007)\\
Station Agent, The (2003)\\
Beginners (2010)
\vfill}
\end{mdframed}
\end{minipage}
\hfill
\begin{minipage}[t]{0.325\textwidth}
\begin{mdframed}[
    linewidth=1pt,
    linecolor=blue!60,
    backgroundcolor=blue!8,
    roundcorner=3pt,
    innertopmargin=2pt,
    innerbottommargin=2pt,
    innerrightmargin=2pt,
    innerleftmargin=2pt,
    skipabove=0pt,
    skipbelow=0pt,
]
\footnotesize
\vbox to 1.2cm{
\centering \textbf{Request}\\
\centering \textit{``Movies with a green ogre''}
\vfill}
\end{mdframed}
\begin{mdframed}[
    linewidth=1pt,
    linecolor=gray!50,
    backgroundcolor=gray!5,
    roundcorner=3pt,
    innertopmargin=2pt,
    innerbottommargin=2pt,
    innerrightmargin=2pt,
    innerleftmargin=2pt,
    skipabove=0pt,
    skipbelow=0pt,
]
\footnotesize
\vbox to 2.4cm{
\centering \textbf{CTRL-Rec Feed}\\
\raggedright Shrek 2 (2004)\\
Shrek (2001)\\
Kung Fu Panda (2008)\\
Jumanji (1995)\\
Hook (1991)
\vfill}
\end{mdframed}
\end{minipage}
\label{fig:qual-examples}
\end{figure}

\subsection{Reachability Experiment} \label{sec:reachability}
In our reachability experiment~\citep{dean_recommendations_2020,curmei21}, we randomly select two users from the MovieLens dataset—a source user and a target user—and evaluate how easily the source user's recommendations can be adjusted to resemble the engagement-based recommendations of the target user using both CTRL-Rec and traditional filters (specifically, for genre and decade). We chose these particular filters to align with the existing controls available on Letterboxd (see \Cref{fig:lb-controls} for a screenshot), a popular movie ratings platform whose users we recruit for our experiments in \Cref{sec:letterboxd}.

Our experimental procedure simulates a user who has a specific target recommendation feed in mind and systematically explores different combinations of filters and LLM requests to achieve it. We compare two approaches: (1) filters only, and (2) filters combined with CTRL-Rec's natural language control. %

\textbf{Feed Quality Evaluation.} For each experiment trial, we generate the top-10 recommendations for both the source and target users using our engagement-based recommender system. We evaluate recommendation quality using two metrics: (1) \textit{cosine similarity distance} (primary), and (2) \textit{percentage overlap} (secondary). The cosine similarity between recommendation feeds is computed as:
\begin{equation}
\text{sim}(F_1, F_2) = \frac{\bar{\mathbf{e}}_{F_1} \cdot \bar{\mathbf{e}}_{F_2}}{\|\bar{\mathbf{e}}_{F_1}\|_2 \|\bar{\mathbf{e}}_{F_2}\|_2}
\end{equation}
where $\bar{\mathbf{e}}_{F} = \frac{1}{|F|}\sum_{i \in F} \mathbf{e}_i$ is the average embedding for feed $F$, and $\mathbf{e}_i$ represents the item embedding for item $i$. This is our primary optimization target as it measures semantic similarity between recommendation sets. However, since embedding distance is approximate and relies on the quality of the underlying embeddings, we also use percentage overlap as a secondary metric, which provides a straightforward ground-truth measure that is easy to interpret -- but is much more noisy, as in many cases the percentage overlap is 0.

\textbf{Filter-Only Approach.} We evaluate various filter combinations (genre and decade) applied to the source user's recommendations. To optimistically estimate users' capacity to use filters effectively, we use a greedy approach based on the target feed's genre and decade distributions. Since movies can belong to many genres and decades, considering all possible filter combinations would be computationally expensive, and most combinations would yield empty feeds with no results. Our greedy approach focuses on filter combinations most likely to yield meaningful results while still aiming to overestimate what users might realistically achieve through manual exploration. 
We select the combination that produces recommendations closest to the target feed in embedding space. See \Cref{app:filter-greedy} for details on our greedy approach.

\textbf{Filter + LLM Approach.} For the CTRL-Rec approach, we simulate an iterative user who refines their natural language requests based on feedback using Gemini Flash 2.5 as the LLM agent. We select starting filter combinations by taking the best-performing filter combination from the filter-only approach and considering all of its subsets as potential starting points. The intuition is that while the best filter combination may be optimal when filters are the only available lever, it might filter out important movies that an LLM could better target through natural language. By allowing the LLM to start from subsets (including no filters at all), we test whether natural language control can compensate for reduced filtering.

For each starting filter combination, an LLM agent: (1) observes current recommendations, (2) compares them to the target feed, (3) generates a natural language request to steer recommendations closer to the target, (4) applies the request via CTRL-Rec, and (5) repeats for up to 3 iterations. The agent refines its strategy based on previous results, simulating how users might iteratively adjust queries. For each starting filter combination, we select the natural language request (across all iterations) that maximizes cosine similarity to the target feed.

\begin{figure}[t]
    \vspace{-1em}
    \centering
    \begin{subfigure}[b]{0.48\textwidth}
        \centering
        \includegraphics[width=\textwidth]{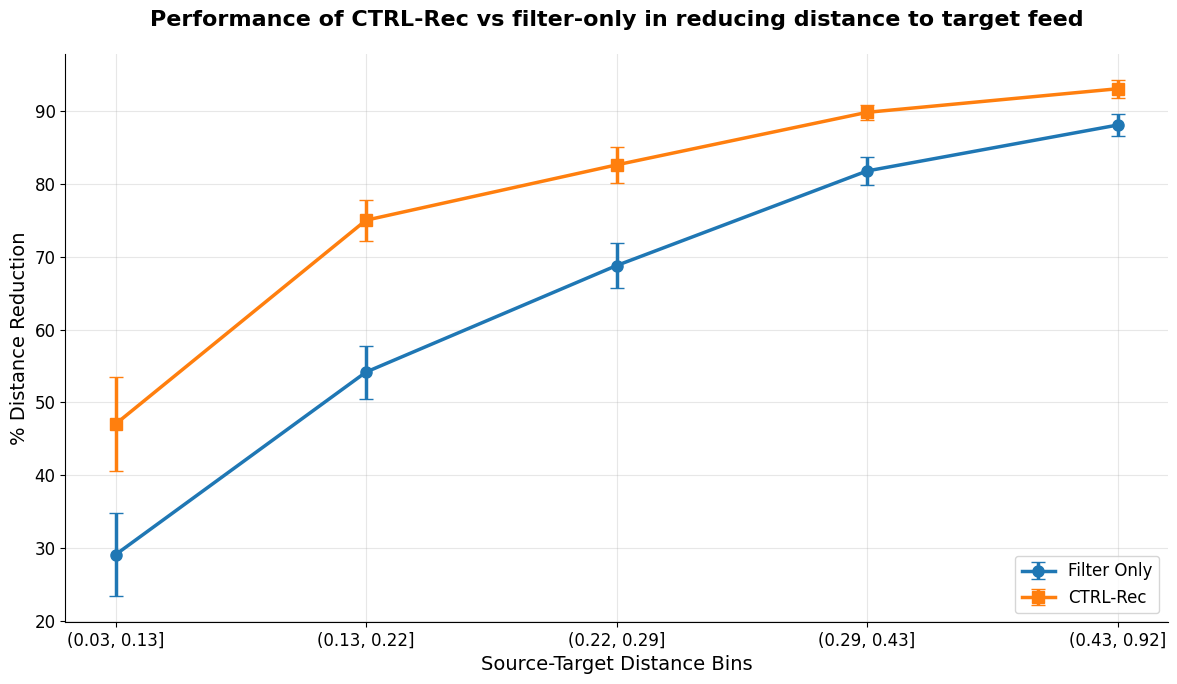}
        \caption{Percentage of distance to target feeds achieved by different approaches.}
        \label{fig:perc-distance}
    \end{subfigure}
    \hfill
    \begin{subfigure}[b]{0.48\textwidth}
        \centering
        \includegraphics[width=\textwidth]{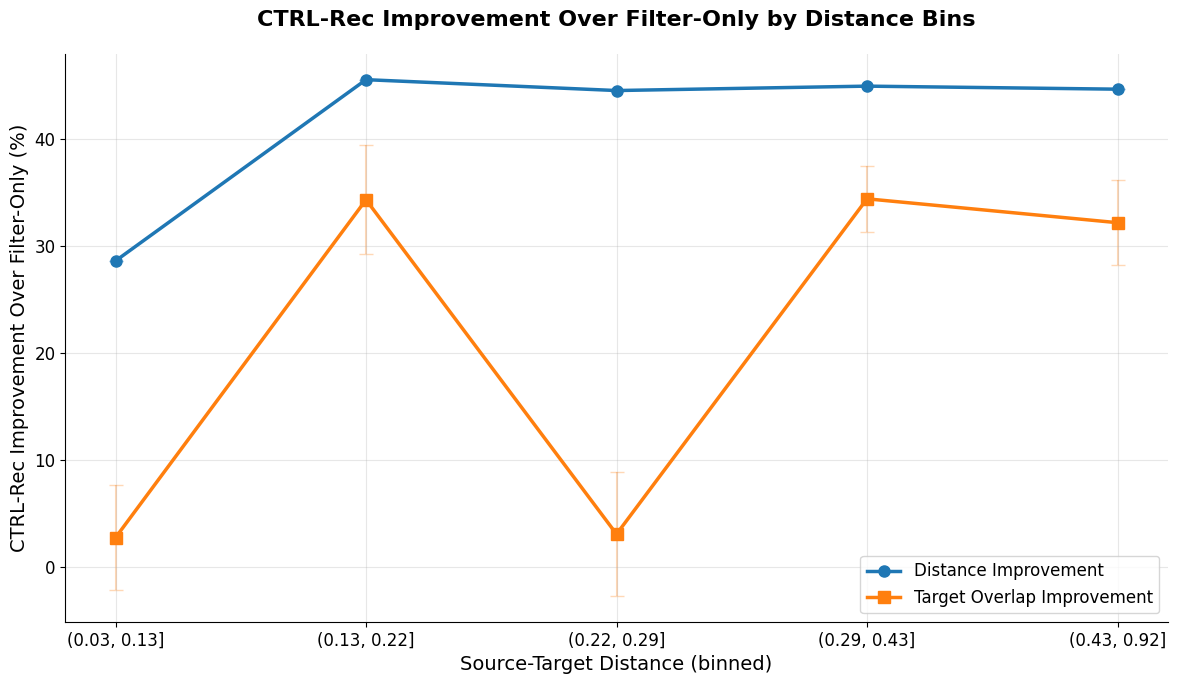}
        \caption{Percentage of the remaining distance to target feed cut by CTRL-Rec over the filter-only approach.}
        \label{fig:improvement-over-filter}
    \end{subfigure}
    \caption{CTRL-Rec reachability experiment results. (a) Traditional filters can already greatly reduce distance to the target feed, but CTRL-Rec meaningfully improves performance further across all distance bins. (b) CTRL-Rec achieves up to $\sim$45\% distance reduction for source-target pairs with higher distance, and 30\% for low distances. The percentage overlap metric shows similar trends, with one outlier showing decreased overlap for high-distance pairs.}
    \label{fig:reachability-main}
    \vspace{-1em}
\end{figure}

\textbf{Results.} As seen in \Cref{fig:reachability-main}, we see that both filters and CTRL-Rec are very effective at bridging the gap between source and target feeds, with a large majority of the gap being closed for source-target pairs which are more distant. However, across all distances, CTRL-Rec consistently outperforms the filters. Metrics are plotted averaged across bins (with standard errors), where bins were formed by taking quintiles of the source-target distances. Moreover, \Cref{fig:reachability-main} also shows that distance reduction in embedding space (our primary measure) is accompanied by an increase in overlap with the target feed (our secondary measure), showing that the embedding distance metric we optimized also correlates with other desirable measures.

\vspace{-0.5em}

\subsection{Human Study with Letterboxd Users} \label{sec:letterboxd}
Our reachability experiment provided insight into the \textit{theoretical} controllability advantages of CTRL-Rec, but we also wanted to see whether and how these advantages surface in \textit{real-world use}. Thus, we conducted a human study with $n=19$ Letterboxd users to evaluate the effectiveness of CTRL-Rec in a movie recommendation task leveraging users' own Letterboxd data. We recruited participants through social media posts, university Slack channels, and random sampling of personal connections that met the study criteria (18+ years old, have 10+ movie ratings on Letterboxd). Our study was approved by our institution's IRB, and our participants were paid a \$20 USD gift card for participating in the 30-minute study.%

\begin{figure}[t!]
    \vspace{-2em}
    \centering
    \includegraphics[width=\textwidth]{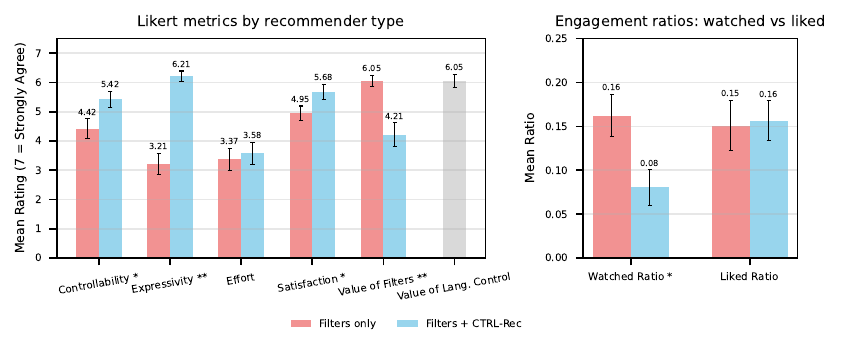}
    \vspace{-2em}
    \caption{Human study results comparing CTRL-Rec with traditional filters to filters-only baseline. Participants found CTRL-Rec's recommendations more engaging, were more satisfied with them, and found them easier to control. Significance levels are indicated using $q$-values (false discovery rate adjusted $p$-values): $^*$ indicates $q<0.05$; $^{**}$ indicates $q<0.01$.}
    \label{fig:human-study}
    \vspace{-1.5em}
\end{figure}

\textbf{Setup.} Our study had a within-subjects design with two counterbalanced conditions: (a) CTRL-Rec\footnote{We used a $w_\text{control}= 0.995$ for the trade-off between engagement and stated preferences (\Cref{eq:realm_score}).} + genre and decade filters, (b) genre and decade filters only. We generated an initial set of recommendations with an engagement-based recommender feed based on their Letterboxd ratings using the recommender in \Cref{sec:reachability}. Participants were then given 8 minutes per condition to use the interface to find their ideal movie recommendations. For each recommended movie, we also provided participants ``\textit{Interested}'' or ``\textit{Watched}'' buttons that they could use to `like' the movie or indicate that they had already seen it. The ``Interested'' (like) button was used to measure user engagement, while the ``\textit{Watched}” button was used to track the rate of redundant recommendations—movies that reflect historical engagement but are unlikely to predict future engagement because the user has already seen them. After each condition, participants answered Likert scale survey questions (on a 7-point scale) to rate their experience with the system. Upon completing both conditions, participants were asked to assess the overall value of the natural language controls and took part in a brief, semi-structured exit interview about their experiences. Further details and study materials can be found in \Cref{app:human-study}.

\textbf{Quantitative Results.} We compared participants' Likert scale ratings across CTRL-Rec and our baseline for a variety of measures, including controllability, satisfaction with recommendations, and engagement. All measures can be found in Table~\ref{t:p-corrected}. To assess statistical significance, we used Wilcoxon signed-rank tests. To account for multiple comparisons, we applied the Benjamini-Krieger-Yekutieli (BKY) two-stage method~\citep{benjamini2006adaptive}, which controls the false discovery rate (FDR). For each measure, we report both the raw $p$-value and the corresponding $q$-value in Table~\ref{t:p-corrected}. The $q$-value represents the minimum FDR at which a result is considered significant; it is an adjusted $p$-value that accounts for the expected proportion of false positives among all significant findings. All results that are significant at the conventional $\alpha = 0.05$ level are also significant at an FDR threshold of $q \leq 0.039$.

Let $M_C$ and $M_B$ denote the mean [Likert rating or engagement ratio] for CTRL-Rec and the baseline, respectively. CTRL-Rec was rated as significantly \textbf{more controllable} than the baseline ($q=0.039; M_C=5.4 > M_B=4.4$). Users could also \textbf{express their preferences significantly more easily} in CTRL-Rec  ($q=0.001; M_C=6.2 > M_B=3.2$). Additionally, participants reported \textbf{greater satisfaction with their recommendations} in CTRL-Rec ($q=0.038; M_C=5.7 > M_B=4.9$). Despite the need to articulate their preferences in natural language when using CTRL-Rec, participants \textbf{did not report a significant increase in effort} compared to the baseline ($q=0.268; M_C=3.6 > M_B=3.4$). Additionally, we found that CTRL-Rec recommended \textbf{significantly fewer movies users have already watched} ($q=0.027; M_C=0.05 < M_B=0.15$), \textbf{but with no significant difference in user engagement}, i.e., the percent of movie recommendations that they `liked' ($q=0.308; M_C=0.16 > M_B=0.14$). Overall, participants \textbf{agreed that the natural language controls in CTRL-Rec were valuable} ($M=6.1 \pm 0.9$).  Once natural language controls were available, participants relied less on the provided filters, and the \textbf{perceived value of filters was significantly lower in CTRL-Rec} compared to the baseline  ($q=0.006; M_C=4.2 < M_B=6.1$). 

\textbf{Qualitative Results.} We qualitatively analyzed participants' quotes from our studies to identify specific factors that made CTRL-Rec more controllable for users. Many participants appreciated that CTRL-Rec allowed them to \textbf{simultaneously find movies based on abstract ``vibes'' while also supporting highly specific queries}. For example, P5 shared that \textit{``I understand the vibe of what I want, but I don't know that I would put it into a particular genre---that's kind of a really broad catch-all.''} P3 thought that questions like \textit{``what's a good film about food to help explain this anthropological problem''} can't be easily answered without natural language controls. The flexibility of CTRL-Rec allowed participants to easily tune the level of specificity: P10 initially wrote a detailed query but then realized ``I don't want to just be given exactly what I want'' and saw value in \textit{``deliberately underspecifying''} her preferences to the system.  %

CTRL-Rec also allowed participants to \textbf{anchor their recommendations on other movies they've enjoyed}. P9 included \textit{``Movies like Dune or Game of Thrones or Lord of the Rings''} in his query. Similarly, P16 described his overall strategy as: \textit{``I was thinking of 5-star movies that I like. And I was trying to find something similar.''} This resulted in recommendations that were \textbf{both high-quality and pleasantly unfamiliar}. Even P20's simple query of \textit{``Find older movies I would like''} yielded recommendations they thought were \textit{``pretty good, I'd never heard of [them] but seems like some things I would like.''} P12 described CTRL-Rec allowed him to reach \textit{``the outer regions of [common recommendations] where there are weird things I haven't seen but interest me,''} whereas the baseline system recommended mostly \textit{``stuff I've heard of, or films that I'd actually watched and forgotten.''} P23 echoed this, sharing that CTRL-Rec's recommendations are hard to come by with conventional search: \textit{``I think the recommendations are really nice. I'm surprised that I don't know about these, so it's exciting. I probably wouldn't have come across them on Google.''} This is also reflected in the quantitative results from \Cref{fig:human-study}: CTRL-Rec generally showed users movies that they were less familiar with, relative to the standard engagement recommender system combined with filters (as seen by ``Watched Ratio'').

Finally, an interesting use case of CTRL-Rec for participants was \textbf{finding recommendations based on group preferences}. P20 queried CTRL-Rec for movies him and his partner could enjoy together, while P23, who organizes regular movie nights with friends, shared that CTRL-Rec would be valuable for finding movies that \textit{``a bunch of friends who are also doing a PhD and who are into sci-fi can really enjoy in a group setting.''}

\section{Discussion}

Our results demonstrate that CTRL-Rec effectively integrates natural language requests into traditional recommender systems, granting users fine-grained real-time control over their recommendations. The success of our distilled approach suggests that natural language control can be made computationally efficient enough for practical deployment, requiring only a single embedding computation per user request. Findings from our human study with Letterboxd users highlight the real-world value of our method in bolstering system controllability and user satisfaction of recommended items without compromising user engagement.

\textbf{Limitations and Future Work.} Our evaluation is limited by the scarcity of publicly available domains that offer both large-scale, up-to-date item inventories and corresponding engagement data. As a result, we focus on the movie recommendation domain, utilizing the MovieLens dataset as our item inventory. While standard LLMs can generate movie recommendations via direct queries, this approach is significantly more computationally expensive. Our chosen domain thus serves as a proof-of-concept for more complex and dynamic settings—such as rapidly evolving social media feeds or large-scale e-commerce catalogs—where the efficiency gains of our method would be especially valuable.

The MovieLens dataset presents its own limitations. It primarily consists of well-known movies that LLMs likely encountered during training, so we do not directly test generalization to truly novel items—a common challenge for retrieval-based systems. Additionally, for our user study, we recruited participants from the Letterboxd platform, primarily due to its ability to export movie ratings that can be used to provide personalized recommendations each user. However, this may introduce bias, as Letterboxd users are typically film enthusiasts and may not reflect the broader population.

Our current evaluation is limited to text-based items, though our framework would naturally extend to other modalities (e.g., images or videos). Finally, we do not address more complex feed-level requests (e.g., “I want half of my recommendations to be of type X, half of type Y”), which remain an interesting avenue for future work.

\textbf{Conclusion.} We have presented CTRL-Rec, a framework for incorporating natural language user controls into traditional recommender systems. Our key insight is that LLMs can be used to simulate how well items align with user requests, and these simulated judgments can be distilled and decomposed into efficient dot products between user-request and item embeddings. This decomposition allows us to reduce the computational cost by orders of magnitude, requiring only one LLM embedding computation per user request rather than per item. Our experiments demonstrate that this approach allows for effective steering of recommendations according to user requests while maintaining high engagement metrics. The method is computationally efficient and practical for real-world deployment. We believe this work represents a step toward recommender systems that better balance engagement optimization with explicit user preferences and control.

\section*{Acknowledgments}
This work has taken place in part in the Rewarding Lab at UT Austin. During this project, the Rewarding Lab has been supported by NSF (IIS-2402650), ONR (N00014-22-1-2204), ARO (W911NF-25-1-0254), Emerson, EA Ventures, UT Austin's Good Systems grand challenge, and Open Philanthropy. We thank Luke Thorburn and Manzil Zaheer for their feedback. We thank the Berkeley Existential Risk Initiative (BERI) for funding our user study.

\newpage

\bibliography{refs}

\begin{thebibliography}{58}
\providecommand{\natexlab}[1]{#1}
\providecommand{\url}[1]{\texttt{#1}}
\expandafter\ifx\csname urlstyle\endcsname\relax
  \providecommand{\doi}[1]{doi: #1}\else
  \providecommand{\doi}{doi: \begingroup \urlstyle{rm}\Url}\fi

\bibitem[Agarwal et~al.(2024)Agarwal, Usunier, Lazaric, and Nickel]{agarwal2024system}
Arpit Agarwal, Nicolas Usunier, Alessandro Lazaric, and Maximilian Nickel.
\newblock System-2 recommenders: Disentangling utility and engagement in recommendation systems via temporal point-processes.
\newblock In \emph{The 2024 ACM Conference on Fairness, Accountability, and Transparency}, pp.\  1763--1773, 2024.

\bibitem[Anderson(2008)]{anderson2008multiple}
Michael~L Anderson.
\newblock Multiple inference and gender differences in the effects of early intervention: A reevaluation of the abecedarian, perry preschool, and early training projects.
\newblock \emph{Journal of the American statistical Association}, 103\penalty0 (484):\penalty0 1481--1495, 2008.

\bibitem[Benjamini et~al.(2006)Benjamini, Krieger, and Yekutieli]{benjamini2006adaptive}
Yoav Benjamini, Abba~M Krieger, and Daniel Yekutieli.
\newblock Adaptive linear step-up procedures that control the false discovery rate.
\newblock \emph{Biometrika}, 93\penalty0 (3):\penalty0 491--507, 2006.

\bibitem[Bonifacio et~al.(2022)Bonifacio, Abonizio, Fadaee, and Nogueira]{bonifacio_inpars_2022}
Luiz Bonifacio, Hugo Abonizio, Marzieh Fadaee, and Rodrigo Nogueira.
\newblock {InPars}: {Data} {Augmentation} for {Information} {Retrieval} using {Large} {Language} {Models}, February 2022.
\newblock URL \url{http://arxiv.org/abs/2202.05144}.
\newblock arXiv:2202.05144 [cs].

\bibitem[Bromley et~al.(1993)Bromley, Guyon, LeCun, S\"{a}ckinger, and Shah]{bromley1993}
Jane Bromley, Isabelle Guyon, Yann LeCun, Eduard S\"{a}ckinger, and Roopak Shah.
\newblock {Signature verification using a "Siamese" time delay neural network}.
\newblock In \emph{Proceedings of the 7th International Conference on Neural Information Processing Systems}, NIPS'93, pp.\  737–744, San Francisco, CA, USA, 1993. Morgan Kaufmann Publishers Inc.

\bibitem[Cunningham et~al.(2025)Cunningham, Pandey, Sigerson, Stray, Allen, Barrilleaux, Iyer, Kothari, Rezaei, Kairam, et~al.]{cunningham2025ranking}
Tom Cunningham, Sana Pandey, Leif Sigerson, Jonathan Stray, Jeff Allen, Bonnie Barrilleaux, Ravi Iyer, Mohit Kothari, Behnam Rezaei, Sanjay Kairam, et~al.
\newblock {Ranking by Engagement and Non-Engagement Signals: Learnings from Industry}.
\newblock \emph{Annals of the New York Academy of Sciences}, 2025.

\bibitem[Curmei et~al.(2021)Curmei, Dean, and Recht]{curmei21}
Mihaela Curmei, Sarah Dean, and Benjamin Recht.
\newblock Quantifying availability and discovery in recommender systems via stochastic reachability.
\newblock In Marina Meila and Tong Zhang (eds.), \emph{Proceedings of the 38th International Conference on Machine Learning}, volume 139 of \emph{Proceedings of Machine Learning Research}, pp.\  2265--2275. PMLR, 18--24 Jul 2021.
\newblock URL \url{https://proceedings.mlr.press/v139/curmei21a.html}.

\bibitem[Dean et~al.(2020)Dean, Rich, and Recht]{dean_recommendations_2020}
Sarah Dean, Sarah Rich, and Benjamin Recht.
\newblock Recommendations and {User} {Agency}: {The} {Reachability} of {Collaboratively}-{Filtered} {Information}.
\newblock In \emph{Proceedings of the 2020 {Conference} on {Fairness}, {Accountability}, and {Transparency}}, pp.\  436--445, January 2020.
\newblock \doi{10.1145/3351095.3372866}.
\newblock URL \url{http://arxiv.org/abs/1912.10068}.
\newblock arXiv:1912.10068 [cs].

\bibitem[Ekstrand \& Willemsen(2016)Ekstrand and Willemsen]{ekstrand2016}
Michael~D. Ekstrand and Martijn~C. Willemsen.
\newblock Behaviorism is not enough: Better recommendations through listening to users.
\newblock In \emph{Proceedings of the 10th ACM Conference on Recommender Systems}, RecSys '16, pp.\  221–224, New York, NY, USA, 2016. Association for Computing Machinery.
\newblock ISBN 9781450340359.
\newblock \doi{10.1145/2959100.2959179}.
\newblock URL \url{https://doi.org/10.1145/2959100.2959179}.

\bibitem[Feng et~al.(2024)Feng, Koo, Tan, Bruckman, McDonald, and Zhang]{feng2024}
K.~J.~Kevin Feng, Xander Koo, Lawrence Tan, Amy Bruckman, David~W. McDonald, and Amy~X. Zhang.
\newblock {Mapping the Design Space of Teachable Social Media Feed Experiences}.
\newblock In \emph{Proceedings of the 2024 CHI Conference on Human Factors in Computing Systems}, CHI '24, New York, NY, USA, 2024. Association for Computing Machinery.
\newblock ISBN 9798400703300.
\newblock \doi{10.1145/3613904.3642120}.
\newblock URL \url{https://doi.org/10.1145/3613904.3642120}.

\bibitem[Friedman et~al.(2023)Friedman, Ahuja, Allen, Tan, Sidahmed, Long, Xie, Schubiner, Patel, Lara, Chu, Chen, and Tiwari]{friedman_leveraging_2023}
Luke Friedman, Sameer Ahuja, David Allen, Zhenning Tan, Hakim Sidahmed, Changbo Long, Jun Xie, Gabriel Schubiner, Ajay Patel, Harsh Lara, Brian Chu, Zexi Chen, and Manoj Tiwari.
\newblock Leveraging {Large} {Language} {Models} in {Conversational} {Recommender} {Systems}, May 2023.
\newblock URL \url{http://arxiv.org/abs/2305.07961}.
\newblock arXiv:2305.07961 [cs].

\bibitem[Gao et~al.(2023)Gao, Sheng, Xiang, Xiong, Wang, and Zhang]{gao_chat-rec_2023}
Yunfan Gao, Tao Sheng, Youlin Xiang, Yun Xiong, Haofen Wang, and Jiawei Zhang.
\newblock Chat-{REC}: {Towards} {Interactive} and {Explainable} {LLMs}-{Augmented} {Recommender} {System}, April 2023.
\newblock URL \url{http://arxiv.org/abs/2303.14524}.
\newblock arXiv:2303.14524 [cs].

\bibitem[Gao et~al.(2025)Gao, Zhou, Dai, and Joachims]{gao_end--end_2025}
Zhaolin Gao, Joyce Zhou, Yijia Dai, and Thorsten Joachims.
\newblock End-to-end {Training} for {Recommendation} with {Language}-based {User} {Profiles}, February 2025.
\newblock URL \url{http://arxiv.org/abs/2410.18870}.
\newblock arXiv:2410.18870 [cs].

\bibitem[Graham et~al.(2019)Graham, Min, and Wu]{graham2019}
Scott Graham, Jun-Ki Min, and Tao Wu.
\newblock {Microsoft Recommenders: Tools to Accelerate Developing Recommender Systems}.
\newblock In \emph{Proceedings of the 13th ACM Conference on Recommender Systems}, RecSys '19, pp.\  542–543, New York, NY, USA, 2019. Association for Computing Machinery.
\newblock ISBN 9781450362436.
\newblock \doi{10.1145/3298689.3346967}.
\newblock URL \url{https://doi.org/10.1145/3298689.3346967}.

\bibitem[He et~al.(2023)He, Xie, Jha, Steck, Liang, Feng, Majumder, Kallus, and McAuley]{he_large_2023}
Zhankui He, Zhouhang Xie, Rahul Jha, Harald Steck, Dawen Liang, Yesu Feng, Bodhisattwa~Prasad Majumder, Nathan Kallus, and Julian McAuley.
\newblock Large {Language} {Models} as {Zero}-{Shot} {Conversational} {Recommenders}.
\newblock In \emph{Proceedings of the 32nd {ACM} {International} {Conference} on {Information} and {Knowledge} {Management}}, pp.\  720--730, October 2023.
\newblock \doi{10.1145/3583780.3614949}.
\newblock URL \url{http://arxiv.org/abs/2308.10053}.
\newblock arXiv:2308.10053 [cs].

\bibitem[Huang \& Chen(2024)Huang and Chen]{huang_pairdistill_2024}
Chao-Wei Huang and Yun-Nung Chen.
\newblock {PairDistill}: {Pairwise} {Relevance} {Distillation} for {Dense} {Retrieval}.
\newblock In Yaser Al-Onaizan, Mohit Bansal, and Yun-Nung Chen (eds.), \emph{Proceedings of the 2024 {Conference} on {Empirical} {Methods} in {Natural} {Language} {Processing}}, pp.\  18225--18237, Miami, Florida, USA, November 2024. Association for Computational Linguistics.
\newblock \doi{10.18653/v1/2024.emnlp-main.1013}.
\newblock URL \url{https://aclanthology.org/2024.emnlp-main.1013/}.

\bibitem[Huang et~al.(2024)Huang, Yu, Xie, Zhang, Yao, and McAuley]{huang_foundation_2024}
Chengkai Huang, Tong Yu, Kaige Xie, Shuai Zhang, Lina Yao, and Julian McAuley.
\newblock Foundation {Models} for {Recommender} {Systems}: {A} {Survey} and {New} {Perspectives}, February 2024.
\newblock URL \url{http://arxiv.org/abs/2402.11143}.
\newblock arXiv:2402.11143 [cs].

\bibitem[Izacard \& Grave(2021)Izacard and Grave]{izacard2020}
Gautier Izacard and Edouard Grave.
\newblock Distilling knowledge from reader to retriever for question answering.
\newblock In \emph{{ICLR}}, 2021.

\bibitem[Jahanbakhsh et~al.(2025)Jahanbakhsh, Zhao, Piccardi, Robertson, Epstein, Koyejo, and Bernstein]{jahanbakhsh2025valuealignmentsocialmedia}
Farnaz Jahanbakhsh, Dora Zhao, Tiziano Piccardi, Zachary Robertson, Ziv Epstein, Sanmi Koyejo, and Michael~S. Bernstein.
\newblock Value alignment of social media ranking algorithms, 2025.
\newblock URL \url{https://arxiv.org/abs/2509.14434}.

\bibitem[Jia et~al.(2024)Jia, Lam, Mai, Hancock, and Bernstein]{jia2024}
Chenyan Jia, Michelle~S. Lam, Minh~Chau Mai, Jeffrey~T. Hancock, and Michael~S. Bernstein.
\newblock {Embedding Democratic Values into Social Media AIs via Societal Objective Functions}.
\newblock \emph{Proc. ACM Hum.-Comput. Interact.}, 8\penalty0 (CSCW1), April 2024.
\newblock \doi{10.1145/3641002}.
\newblock URL \url{https://doi.org/10.1145/3641002}.

\bibitem[Karpukhin et~al.(2020)Karpukhin, Oğuz, Min, Lewis, Wu, Edunov, Chen, and Yih]{karpukhin_dense_2020}
Vladimir Karpukhin, Barlas Oğuz, Sewon Min, Patrick Lewis, Ledell Wu, Sergey Edunov, Danqi Chen, and Wen-tau Yih.
\newblock Dense {Passage} {Retrieval} for {Open}-{Domain} {Question} {Answering}, September 2020.
\newblock URL \url{http://arxiv.org/abs/2004.04906}.
\newblock arXiv:2004.04906 [cs].

\bibitem[Khattab \& Zaharia(2020)Khattab and Zaharia]{khattab2020}
Omar Khattab and Matei Zaharia.
\newblock {ColBERT: Efficient and Effective Passage Search via Contextualized Late Interaction over BERT}.
\newblock In \emph{Proceedings of the 43rd International ACM SIGIR Conference on Research and Development in Information Retrieval}, SIGIR '20, pp.\  39–48, New York, NY, USA, 2020. Association for Computing Machinery.
\newblock ISBN 9781450380164.
\newblock \doi{10.1145/3397271.3401075}.
\newblock URL \url{https://doi.org/10.1145/3397271.3401075}.

\bibitem[Kim et~al.(2024)Kim, Kim, Cho, Kang, Chang, Yeo, and Lee]{kim2024reviewdrivenpersonalizedpreferencereasoning}
Jieyong Kim, Hyunseo Kim, Hyunjin Cho, SeongKu Kang, Buru Chang, Jinyoung Yeo, and Dongha Lee.
\newblock Review-driven personalized preference reasoning with large language models for recommendation, 2024.
\newblock URL \url{https://arxiv.org/abs/2408.06276}.

\bibitem[Kleinberg et~al.(2024{\natexlab{a}})Kleinberg, Ludwig, Mullainathan, and Raghavan]{kleinberg2024inversion}
Jon Kleinberg, Jens Ludwig, Sendhil Mullainathan, and Manish Raghavan.
\newblock The inversion problem: Why algorithms should infer mental state and not just predict behavior.
\newblock \emph{Perspectives on Psychological Science}, 19\penalty0 (5):\penalty0 827--838, 2024{\natexlab{a}}.

\bibitem[Kleinberg et~al.(2024{\natexlab{b}})Kleinberg, Mullainathan, and Raghavan]{kleinberg2024challenge}
Jon Kleinberg, Sendhil Mullainathan, and Manish Raghavan.
\newblock {The challenge of understanding what users want: Inconsistent preferences and engagement optimization}.
\newblock \emph{Management science}, 70\penalty0 (9):\penalty0 6336--6355, 2024{\natexlab{b}}.

\bibitem[Kolluri et~al.(2026)Kolluri, Su, Jahanbakhsh, Zhao, Piccardi, and Bernstein]{kolluri2026alexandrialibrarypluralisticvalues}
Akaash Kolluri, Renn Su, Farnaz Jahanbakhsh, Dora Zhao, Tiziano Piccardi, and Michael~S. Bernstein.
\newblock {Alexandria: A Library of Pluralistic Values for Realtime Re-Ranking of Social Media Feeds}.
\newblock In \emph{Proceedings of the International Conference on Web and Social Media (ICWSM)}, 2026.

\bibitem[Lazar et~al.(2024)Lazar, Thorburn, Jin, and Belli]{lazar_moral_2024}
Seth Lazar, Luke Thorburn, Tian Jin, and Luca Belli.
\newblock The {Moral} {Case} for {Using} {Language} {Model} {Agents} for {Recommendation}, October 2024.
\newblock URL \url{http://arxiv.org/abs/2410.12123}.
\newblock arXiv:2410.12123 [cs].

\bibitem[Li et~al.(2025)Li, Jin, Zhou, Zhang, Zhang, Zhu, and Dou]{li2025}
Xiaoxi Li, Jiajie Jin, Yujia Zhou, Yuyao Zhang, Peitian Zhang, Yutao Zhu, and Zhicheng Dou.
\newblock {From Matching to Generation: A Survey on Generative Information Retrieval}.
\newblock \emph{ACM Trans. Inf. Syst.}, 43\penalty0 (3), May 2025.
\newblock ISSN 1046-8188.
\newblock \doi{10.1145/3722552}.
\newblock URL \url{https://doi.org/10.1145/3722552}.

\bibitem[Lin et~al.(2024)Lin, Dai, Xi, Liu, Chen, Zhang, Liu, Wu, Li, Zhu, Guo, Yu, Tang, and Zhang]{lin_how_2024}
Jianghao Lin, Xinyi Dai, Yunjia Xi, Weiwen Liu, Bo~Chen, Hao Zhang, Yong Liu, Chuhan Wu, Xiangyang Li, Chenxu Zhu, Huifeng Guo, Yong Yu, Ruiming Tang, and Weinan Zhang.
\newblock How {Can} {Recommender} {Systems} {Benefit} from {Large} {Language} {Models}: {A} {Survey}, July 2024.
\newblock URL \url{http://arxiv.org/abs/2306.05817}.
\newblock arXiv:2306.05817 [cs].

\bibitem[Lukoff et~al.(2021)Lukoff, Lyngs, Zade, Liao, Choi, Fan, Munson, and Hiniker]{lukoff2021}
Kai Lukoff, Ulrik Lyngs, Himanshu Zade, J.~Vera Liao, James Choi, Kaiyue Fan, Sean~A. Munson, and Alexis Hiniker.
\newblock {How the Design of YouTube Influences User Sense of Agency}.
\newblock In \emph{Proceedings of the 2021 CHI Conference on Human Factors in Computing Systems}, CHI '21, New York, NY, USA, 2021. Association for Computing Machinery.
\newblock ISBN 9781450380966.
\newblock \doi{10.1145/3411764.3445467}.
\newblock URL \url{https://doi.org/10.1145/3411764.3445467}.

\bibitem[Lukoff et~al.(2023)Lukoff, Lyngs, Shirokova, Rao, Tian, Zade, Munson, and Hiniker]{lukoff2023}
Kai Lukoff, Ulrik Lyngs, Karina Shirokova, Raveena Rao, Larry Tian, Himanshu Zade, Sean~A. Munson, and Alexis Hiniker.
\newblock {SwitchTube: A Proof-of-Concept System Introducing “Adaptable Commitment Interfaces” as a Tool for Digital Wellbeing}.
\newblock In \emph{Proceedings of the 2023 CHI Conference on Human Factors in Computing Systems}, CHI '23, New York, NY, USA, 2023. Association for Computing Machinery.
\newblock ISBN 9781450394215.
\newblock \doi{10.1145/3544548.3580703}.
\newblock URL \url{https://doi.org/10.1145/3544548.3580703}.

\bibitem[Malki et~al.(2025)Malki, Quéré, Monroy-Hernández, and Ribeiro]{malki2025bonsaiintentionalpersonalizedsocial}
Omar~El Malki, Marianne Aubin~Le Quéré, Andrés Monroy-Hernández, and Manoel~Horta Ribeiro.
\newblock {Bonsai: Intentional and Personalized Social Media Feeds}, 2025.
\newblock URL \url{https://arxiv.org/abs/2509.10776}.

\bibitem[Milli et~al.(2021)Milli, Belli, and Hardt]{milli2021optimizing}
Smitha Milli, Luca Belli, and Moritz Hardt.
\newblock From {O}ptimizing {E}ngagement to {M}easuring {V}alue.
\newblock In \emph{Proceedings of the 2021 ACM Conference on Fairness, Accountability, and Transparency}, pp.\  714--722, 2021.

\bibitem[Milli et~al.(2023)Milli, Pierson, and Garg]{milli2023choosing}
Smitha Milli, Emma Pierson, and Nikhil Garg.
\newblock Choosing the right weights: Balancing value, strategy, and noise in recommender systems.
\newblock \emph{arXiv preprint arXiv:2305.17428}, 2023.

\bibitem[Milli et~al.(2025)Milli, Carroll, Wang, Pandey, Zhao, and Dragan]{milli2025engagementusersatisfactionamplification}
Smitha Milli, Micah Carroll, Yike Wang, Sashrika Pandey, Sebastian Zhao, and Anca~D Dragan.
\newblock Engagement, user satisfaction, and the amplification of divisive content on social media.
\newblock \emph{PNAS Nexus}, 4\penalty0 (3):\penalty0 pgaf062, 03 2025.
\newblock ISSN 2752-6542.
\newblock \doi{10.1093/pnasnexus/pgaf062}.
\newblock URL \url{https://doi.org/10.1093/pnasnexus/pgaf062}.

\bibitem[Morewedge et~al.(2023)Morewedge, Mullainathan, Naushan, Sunstein, Kleinberg, Raghavan, and Ludwig]{morewedge2023human}
Carey~K Morewedge, Sendhil Mullainathan, Haaya~F Naushan, Cass~R Sunstein, Jon Kleinberg, Manish Raghavan, and Jens~O Ludwig.
\newblock {Human bias in algorithm design}.
\newblock \emph{Nature Human Behaviour}, 7\penalty0 (11):\penalty0 1822--1824, 2023.

\bibitem[Mozilla(2023)]{mozilla_youtube_regrets}
Mozilla.
\newblock Youtube regrets: A crowdsourced investigation into youtube's recommendation algorithm.
\newblock \url{https://foundation.mozilla.org/en/youtube/findings/}, 2023.
\newblock Accessed: 2023-10-17.

\bibitem[Mysore et~al.(2023{\natexlab{a}})Mysore, Jasim, McCallum, and Zamani]{mysore_editable_2023}
Sheshera Mysore, Mahmood Jasim, Andrew McCallum, and Hamed Zamani.
\newblock Editable {User} {Profiles} for {Controllable} {Text} {Recommendation}, October 2023{\natexlab{a}}.
\newblock URL \url{http://arxiv.org/abs/2304.04250}.
\newblock arXiv:2304.04250 [cs].

\bibitem[Mysore et~al.(2023{\natexlab{b}})Mysore, Mccallum, and Zamani]{mysore2023}
Sheshera Mysore, Andrew Mccallum, and Hamed Zamani.
\newblock Large language model augmented narrative driven recommendations.
\newblock In \emph{Proceedings of the 17th ACM Conference on Recommender Systems}, RecSys '23, pp.\  777–783, New York, NY, USA, 2023{\natexlab{b}}. Association for Computing Machinery.
\newblock ISBN 9798400702419.
\newblock \doi{10.1145/3604915.3608829}.
\newblock URL \url{https://doi.org/10.1145/3604915.3608829}.

\bibitem[Paischer et~al.(2024)Paischer, Yang, Liu, Shao, Hassani, Li, Chen, Li, Gao, Shao, Feng, Noorshams, Park, Long, and Eghbalzadeh]{paischer_preference_2024}
Fabian Paischer, Liu Yang, Linfeng Liu, Shuai Shao, Kaveh Hassani, Jiacheng Li, Ricky Chen, Zhang~Gabriel Li, Xialo Gao, Wei Shao, Xue Feng, Nima Noorshams, Sem Park, Bo~Long, and Hamid Eghbalzadeh.
\newblock Preference {Discerning} with {LLM}-{Enhanced} {Generative} {Retrieval}, December 2024.
\newblock URL \url{http://arxiv.org/abs/2412.08604}.
\newblock arXiv:2412.08604 [cs].

\bibitem[Penaloza et~al.(2025)Penaloza, Gouvert, Wu, and Charlin]{penaloza_tears_2025}
Emiliano Penaloza, Olivier Gouvert, Haolun Wu, and Laurent Charlin.
\newblock {TEARS}: {Textual} {Representations} for {Scrutable} {Recommendations}, March 2025.
\newblock URL \url{http://arxiv.org/abs/2410.19302}.
\newblock arXiv:2410.19302 [cs].

\bibitem[Piccardi et~al.(2024)Piccardi, Saveski, Jia, Hancock, Tsai, and Bernstein]{piccardi2024social}
Tiziano Piccardi, Martin Saveski, Chenyan Jia, Jeffrey~T Hancock, Jeanne~L Tsai, and Michael Bernstein.
\newblock Social media algorithms can shape affective polarization via exposure to antidemocratic attitudes and partisan animosity.
\newblock \emph{arXiv preprint arXiv:2411.14652}, 2024.

\bibitem[Pradeep et~al.(2023)Pradeep, Hui, Gupta, Lelkes, Zhuang, Lin, Metzler, and Tran]{pradeep-etal-2023-generative}
Ronak Pradeep, Kai Hui, Jai Gupta, Adam Lelkes, Honglei Zhuang, Jimmy Lin, Donald Metzler, and Vinh Tran.
\newblock How does generative retrieval scale to millions of passages?
\newblock In Houda Bouamor, Juan Pino, and Kalika Bali (eds.), \emph{Proceedings of the 2023 Conference on Empirical Methods in Natural Language Processing}, pp.\  1305--1321, Singapore, December 2023. Association for Computational Linguistics.
\newblock \doi{10.18653/v1/2023.emnlp-main.83}.
\newblock URL \url{https://aclanthology.org/2023.emnlp-main.83/}.

\bibitem[Radlinski et~al.(2022)Radlinski, Balog, Diaz, Dixon, and Wedin]{radlinski_natural_2022}
Filip Radlinski, Krisztian Balog, Fernando Diaz, Lucas Dixon, and Ben Wedin.
\newblock On {Natural} {Language} {User} {Profiles} for {Transparent} and {Scrutable} {Recommendation}.
\newblock In \emph{Proceedings of the 45th {International} {ACM} {SIGIR} {Conference} on {Research} and {Development} in {Information} {Retrieval}}, pp.\  2863--2874, July 2022.
\newblock \doi{10.1145/3477495.3531873}.
\newblock URL \url{http://arxiv.org/abs/2205.09403}.
\newblock arXiv:2205.09403 [cs].

\bibitem[Rajput et~al.(2023)Rajput, Mehta, Singh, Hulikal~Keshavan, Vu, Heldt, Hong, Tay, Tran, Samost, Kula, Chi, and Sathiamoorthy]{rajput2023}
Shashank Rajput, Nikhil Mehta, Anima Singh, Raghunandan Hulikal~Keshavan, Trung Vu, Lukasz Heldt, Lichan Hong, Yi~Tay, Vinh Tran, Jonah Samost, Maciej Kula, Ed~Chi, and Maheswaran Sathiamoorthy.
\newblock {Recommender Systems with Generative Retrieval}.
\newblock In A.~Oh, T.~Naumann, A.~Globerson, K.~Saenko, M.~Hardt, and S.~Levine (eds.), \emph{Advances in Neural Information Processing Systems}, volume~36, pp.\  10299--10315. Curran Associates, Inc., 2023.
\newblock URL \url{https://proceedings.neurips.cc/paper_files/paper/2023/file/20dcab0f14046a5c6b02b61da9f13229-Paper-Conference.pdf}.

\bibitem[Ramos et~al.(2024)Ramos, Rahmani, Wang, Fu, and Lipani]{ramos_transparent_2024}
Jerome Ramos, Hossen~A. Rahmani, Xi~Wang, Xiao Fu, and Aldo Lipani.
\newblock Transparent and {Scrutable} {Recommendations} {Using} {Natural} {Language} {User} {Profiles}, July 2024.
\newblock URL \url{http://arxiv.org/abs/2402.05810}.
\newblock arXiv:2402.05810 [cs].

\bibitem[Rathje et~al.(2024)Rathje, Robertson, Brady, and Van~Bavel]{rathje2024people}
Steve Rathje, Claire Robertson, William~J Brady, and Jay~J Van~Bavel.
\newblock {People think that social media platforms do (but should not) amplify divisive content}.
\newblock \emph{Perspectives on Psychological Science}, 19\penalty0 (5):\penalty0 781--795, 2024.

\bibitem[Rezk et~al.(2024)Rezk, Simkute, Luger, Vines, Elsden, Evans, and Jones]{rezk2024}
Anna~Marie Rezk, Auste Simkute, Ewa Luger, John Vines, Chris Elsden, Michael Evans, and Rhianne Jones.
\newblock {Agency Aspirations: Understanding Users' Preferences And Perceptions Of Their Role In Personalised News Curation}.
\newblock In \emph{Proceedings of the 2024 CHI Conference on Human Factors in Computing Systems}, CHI '24, New York, NY, USA, 2024. Association for Computing Machinery.
\newblock ISBN 9798400703300.
\newblock \doi{10.1145/3613904.3642634}.
\newblock URL \url{https://doi.org/10.1145/3613904.3642634}.

\bibitem[Sanner et~al.(2023)Sanner, Balog, Radlinski, Wedin, and Dixon]{sanner_large_2023}
Scott Sanner, Krisztian Balog, Filip Radlinski, Ben Wedin, and Lucas Dixon.
\newblock Large {Language} {Models} are {Competitive} {Near} {Cold}-start {Recommenders} for {Language}- and {Item}-based {Preferences}, July 2023.
\newblock URL \url{http://arxiv.org/abs/2307.14225}.
\newblock arXiv:2307.14225 [cs].

\bibitem[Smith(2021)]{smith_2021}
Ben Smith.
\newblock How {T}ik{T}ok reads your mind, Dec 2021.
\newblock URL \url{https://www.nytimes.com/2021/12/05/business/media/tiktok-algorithm.html}.

\bibitem[Tan et~al.(2025)Tan, Ram, Messerschmidt, Dissanayake, and Nanayakkara]{fangyi2025}
Felicia Fang-Yi Tan, Ashwin Ram, Moritz~Alexander Messerschmidt, Hasini~Amanda Dissanayake, and Suranga Nanayakkara.
\newblock Curious shorts: Curiosity-driven exploration and learning on short-form video platforms.
\newblock In \emph{Proceedings of the 2025 CHI Conference on Human Factors in Computing Systems}, CHI '25, New York, NY, USA, 2025. Association for Computing Machinery.
\newblock ISBN 9798400713941.
\newblock \doi{10.1145/3706598.3713951}.
\newblock URL \url{https://doi.org/10.1145/3706598.3713951}.

\bibitem[Twitter(2023)]{Twitter_2023}
Twitter.
\newblock Twitter's {R}ecommendation {A}lgorithm - {H}eavy {R}anker and {T}w{HIN} embeddings, Mar 2023.
\newblock URL \url{https://github.com/twitter/the-algorithm-ml/tree/main/projects/home/recap}.

\bibitem[Wang et~al.(2024)Wang, Li, Wang, Xing, Niu, Kong, Li, Long, Chang, and Zhang]{wang_towards_2024}
Qi~Wang, Jindong Li, Shiqi Wang, Qianli Xing, Runliang Niu, He~Kong, Rui Li, Guodong Long, Yi~Chang, and Chengqi Zhang.
\newblock Towards {Next}-{Generation} {LLM}-based {Recommender} {Systems}: {A} {Survey} and {Beyond}, October 2024.
\newblock URL \url{http://arxiv.org/abs/2410.19744}.
\newblock arXiv:2410.19744 [cs].

\bibitem[Williams et~al.(2024)Williams, Carroll, Narang, Weisser, Murphy, and Dragan]{williams_targeted_2024}
Marcus Williams, Micah Carroll, Adhyyan Narang, Constantin Weisser, Brendan Murphy, and Anca Dragan.
\newblock On {Targeted} {Manipulation} and {Deception} when {Optimizing} {LLMs} for {User} {Feedback}, November 2024.
\newblock URL \url{http://arxiv.org/abs/2411.02306}.
\newblock arXiv:2411.02306 [cs].

\bibitem[Wu et~al.(2020)Wu, Qiao, Chen, Wu, Qi, Lian, Liu, Xie, Gao, Wu, and Zhou]{wu-etal-2020-mind}
Fangzhao Wu, Ying Qiao, Jiun-Hung Chen, Chuhan Wu, Tao Qi, Jianxun Lian, Danyang Liu, Xing Xie, Jianfeng Gao, Winnie Wu, and Ming Zhou.
\newblock {{MIND}: A Large-scale Dataset for News Recommendation}.
\newblock In Dan Jurafsky, Joyce Chai, Natalie Schluter, and Joel Tetreault (eds.), \emph{Proceedings of the 58th Annual Meeting of the Association for Computational Linguistics}, pp.\  3597--3606, Online, July 2020. Association for Computational Linguistics.
\newblock \doi{10.18653/v1/2020.acl-main.331}.
\newblock URL \url{https://aclanthology.org/2020.acl-main.331/}.

\bibitem[Wu et~al.(2024)Wu, Zheng, Qiu, Wang, Gu, Shen, Qin, Zhu, Zhu, Liu, Xiong, and Chen]{wu_survey_2024}
Likang Wu, Zhi Zheng, Zhaopeng Qiu, Hao Wang, Hongchao Gu, Tingjia Shen, Chuan Qin, Chen Zhu, Hengshu Zhu, Qi~Liu, Hui Xiong, and Enhong Chen.
\newblock A {Survey} on {Large} {Language} {Models} for {Recommendation}, June 2024.
\newblock URL \url{http://arxiv.org/abs/2305.19860}.
\newblock arXiv:2305.19860 [cs].

\bibitem[Yang et~al.(2019)Yang, Sobolev, Wang, Chen, Dunne, Tsangouri, Dell, Naaman, and Estrin]{yang2019}
Longqi Yang, Michael Sobolev, Yu~Wang, Jenny Chen, Drew Dunne, Christina Tsangouri, Nicola Dell, Mor Naaman, and Deborah Estrin.
\newblock How intention informed recommendations modulate choices: A field study of spoken word content.
\newblock In \emph{The World Wide Web Conference}, WWW '19, pp.\  2169–2180, New York, NY, USA, 2019. Association for Computing Machinery.
\newblock ISBN 9781450366748.
\newblock \doi{10.1145/3308558.3313540}.
\newblock URL \url{https://doi.org/10.1145/3308558.3313540}.

\bibitem[Zhang et~al.(2022)Zhang, Lukoff, Rao, Baughan, and Hiniker]{zhang2022}
Mingrui~Ray Zhang, Kai Lukoff, Raveena Rao, Amanda Baughan, and Alexis Hiniker.
\newblock {Monitoring Screen Time or Redesigning It? Two Approaches to Supporting Intentional Social Media Use}.
\newblock In \emph{Proceedings of the 2022 CHI Conference on Human Factors in Computing Systems}, CHI '22, New York, NY, USA, 2022. Association for Computing Machinery.
\newblock ISBN 9781450391573.
\newblock \doi{10.1145/3491102.3517722}.
\newblock URL \url{https://doi.org/10.1145/3491102.3517722}.

\end{thebibliography}
\bibliographystyle{iclr2026_conference}

\newpage
\appendix

\section{Usage of LLMs.}
We used LLMs to help with rephrasing individual paragraphs of the paper for improved clarity, and writing code for the experiments.

\section{Prompts}\label{appendix:prompts}

\subsection{Movie Summary Generation}
To ensure consistent and informative movie descriptions, we pre-generate summaries for all movies using the following prompt:

\begin{mdframed}[style=MyFrame]
\small
\textbf{System:} You are a knowledgeable film critic. Provide accurate movie summaries.

\textbf{User:} You are tasked with generating a summary of a specific movie. This summary should be maximally helpful for someone deciding whether to recommend the movie to another person based on their preferences. Follow these instructions carefully:

You will be given the following information:

\texttt{<movie\_title>}\\
\{movie\_title\}\\
\texttt{</movie\_title>}

Using this information, create a comprehensive summary of the movie. Your summary should focus on aspects that would be most relevant when considering whether to recommend the movie to someone. Include the following elements:

\begin{enumerate}[leftmargin=*]
    \item Basic information: Briefly mention the director, main cast, and genre
    \item Plot overview: Provide a high-level summary of the plot including spoilers
    \item Themes and tone: Describe the main themes explored in the movie and its overall tone (e.g., dark, lighthearted, thought-provoking)
    \item Cinematic elements: Highlight notable aspects of cinematography, music, or special effects if they are particularly significant
    \item Critical reception: Mention how the movie was generally received by critics and audiences
    \item Potential appeal: Describe the types of viewers who might enjoy this movie (e.g., fans of certain genres, people interested in specific themes)
    \item Content advisories: Mention any content that might make the movie unsuitable for certain audiences (e.g., violence, sexual content, complex themes)
\end{enumerate}

Keep your summary focused and relevant. Avoid unnecessary details or trivia that wouldn't help in deciding whether to recommend the movie.
\end{mdframed}

\subsection{Genre User Request Generation}\label{app:genre-prompt}
For generating user requests focused on specific genres, we use the following prompt:

\begin{mdframed}[style=MyFrame]
\small
\textbf{System:} You are an AI generating natural language commands that users might give to a movie recommendation system. You should generate a brief first-person statement about which single movie genre the user wants to see right now. Only mention one genre they want to see (not ones to avoid) and only use genres from this specific list: Drama, Comedy, Romance, Thriller, Action, Crime, Adventure, Children, Mystery, Sci-Fi, Fantasy, Horror. The genre requested should reflect what you think this user would genuinely want to watch next based on their rating history.

\textbf{User:} Consider the following movie ratings history for the user:

\texttt{<movie\_ratings\_history>}\\
\{movie\_ratings\_str\}\\
\texttt{</movie\_ratings\_history>}

Guidelines for the statement:
\begin{itemize}[leftmargin=*]
\item Write in first-person perspective
\item Do not mention specific movies or ratings
\item Only mention one genre you want to see (not ones you want to avoid)
\item Only use genres from the provided list
\item The genre you request should be representative of what you would likely want to watch next, based on your rating history
\item Aim for a natural, conversational tone
\item Should be a single sentence mentioning exactly one genre
\item Do not mention any other movie aspects besides this specific genre
\end{itemize}
\end{mdframed}   

\subsection{Open-Ended User Request Generation}\label{app:open-ended-prompt}

For generating more complex, open-ended user requests, we use the following prompt:

\begin{mdframed}[style=MyFrame]
\small
\textbf{System:} You are an AI generating diverse natural language commands that users might give to a movie recommendation system. You should generate a brief first-person statement about which kinds of movies the user wants to see right now.

\textbf{User:} Guidelines for the statement:
\begin{itemize}[leftmargin=*]
\item Write in first-person perspective
\item Do not mention specific movies or ratings
\item Focus on preferences, likes, and dislikes related to movies
\item The user may express interest in movies based on any aspect, such as genres, themes, storytelling styles, visual elements, acting, and production quality
\item Aim for a natural, conversational, and informal tone, as if you were quickly expressing your preferences. Should almost always be a single sentence, potentially even a very short one
\item Vary the structure and focus of the statement to add diversity
\end{itemize}

Consider the following movie ratings history for the user:

\texttt{<movie\_ratings\_history>}\\
\{movie\_ratings\_str\}\\
\texttt{</movie\_ratings\_history>}

Provide the final statement within \texttt{<statement>} tags. Here is an example output structure (do not copy the content, only the format):

\texttt{<statement>}\\
Final first-person statement about what kind of movie the user is currently looking for\\
\texttt{</statement>}

\{previous\_requests\}

Be creative with the style of requests. Some examples of different styles:
\begin{itemize}[leftmargin=*]
\item Direct and simple (``I want something funny'')
\item Descriptive (``Looking for an emotional drama that will make me think'')
\item Mood-based (``I'm in the mood for something thrilling and suspenseful'')
\item Preference-focused (``I prefer movies with complex characters and deep themes'')
\item Contextual (``Need a light comedy for a relaxing evening'')
\item Comparative (``Want something like my favorite action movies but with more humor'')
\end{itemize}

MAKE SURE you open and close your statement tags correctly, \texttt{<statement>} and \texttt{</statement>}.
\end{mdframed}

\subsection{Distilled Model User Request Generation}

For generating diverse training data for the distilled model, we use the following prompt:

\begin{mdframed}[style=MyFrame]
\small
\textbf{System:} You are an AI generating diverse natural language commands that users might give to a movie recommendation system. The statements can include mixes of genres, or say they want to avoid certain genres, or both.

\textbf{User:} Here are some previous requests you've generated. Try to keep them diverse and focus on areas you think you may have missed.

\texttt{<jsonl>}\\
\{requests\_str\}\\
\texttt{</jsonl>}

Please respond only with a JSONL list of 20 new request strings in a similar format, continuing the id counter from the previous requests. Do not output more than 20 new requests (stop at id 40). MAKE SURE to format the JSONL correctly.
Write your response in \texttt{<jsonl>}\texttt{</jsonl>} tags.
\end{mdframed}

\subsection{Feed-level LLM Judge Prompt}

For evaluating how well a set of recommendations matches a user's request, we use the following prompt:

\begin{mdframed}[style=MyFrame]
\small
\textbf{System:} You are an expert movie recommendation system evaluator. Your task is to rate how well a list of recommended movies matches a user's request.

\textbf{User:} Consider a user with the following movie request:

\texttt{<user\_preferences>}\\
\{user\_preference\_text\}\\
\texttt{</user\_preferences>}

Here are the top \{top\_k\} movie recommendations for this user:\\
\{recommendations\_str\}

Rate how well these recommendations match the user's request on a scale of 1-5:
\begin{itemize}[leftmargin=*]
\item 1 = Poor match, recommendations don't reflect user request
\item 3 = Good match, many recommendations align with request  
\item 5 = Excellent match, all recommendations strongly align with request
\end{itemize}

Answer only with an integer from 1 to 5. Do not include any other text.

Answer:
\end{mdframed}

\section{Extracting Scores from LLM}\label{appendix:llm_scoring}

To obtain more calibrated granular scores from LLM judgements, we follow the methodology of \citet{williams_targeted_2024}. Rather than directly requesting a single rating on a scale of 1-5, we leverage the model's underlying probability distribution. Specifically, we extract the logprobs for tokens ``1" through ``5", normalize them into a probability distribution $P(r)$, and compute the expected rating as $\sum_{r=1}^5 r \cdot P(r)$. To verify model comprehension, we ensure the total probability mass on these five tokens exceeds 0.9.

\subsection{Computational Efficiency and Quality of Distillation}\label{subsec:efficiency}

Before considering the results of the experiments described in \Cref{sec:exps}, we discuss the efficiency and quality of our trained distilled model.

\begin{figure}[t]
    \vspace{-1.5em}
    \centering
    \includegraphics[width=1\linewidth]{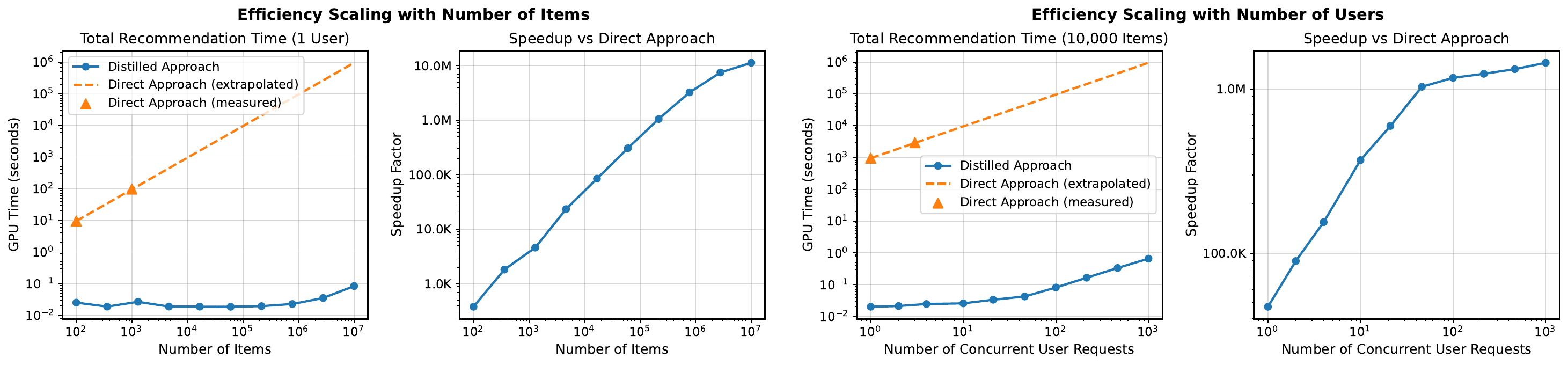}
    \vspace{-1.5em}
    \caption{\textbf{Computational efficiency: the distilled approach fares far better than the direct approach at scale.} The distilled approach requires many orders of magnitude less computational resources than the direct approach for generating recommendations. Moreover, the speed-up factor for the distilled approach grows both with the number of candidate items and concurrent user requests. With $10^7$ items and $1000$ concurrent user requests, \textit{we estimate that the distilled approach would likely be at least $10^8$ times more efficient than direct LLM scoring}.
    All evaluations were done on a single NVIDIA RTX A6000 GPU.}
    \label{fig:item_scaling_benchmark}
    \vspace{-1em}
\end{figure}

\textbf{Distilled Model Computational Efficiency.} We benchmarked the computational efficiency of the distilled approach compared to the direct LLM scoring approach. As seen in \Cref{fig:item_scaling_benchmark}, the computational resources required for the direct approach grow linearly with the number of candidate items and the number of concurrent user requests. Given the high starting cost even for small item sets and a single user, it is clearly infeasible to deploy at scale without massive computational resources and parallelization. While we expect the distilled approach will also grow linearly at large scales, it's far lower starting cost makes it much more applicable to real-world recommendation scenarios.

\section{Reachability Experiment Details}\label{app:reachability}

\begin{figure}[h!]
    \centering
    \includegraphics[width=0.8\textwidth]{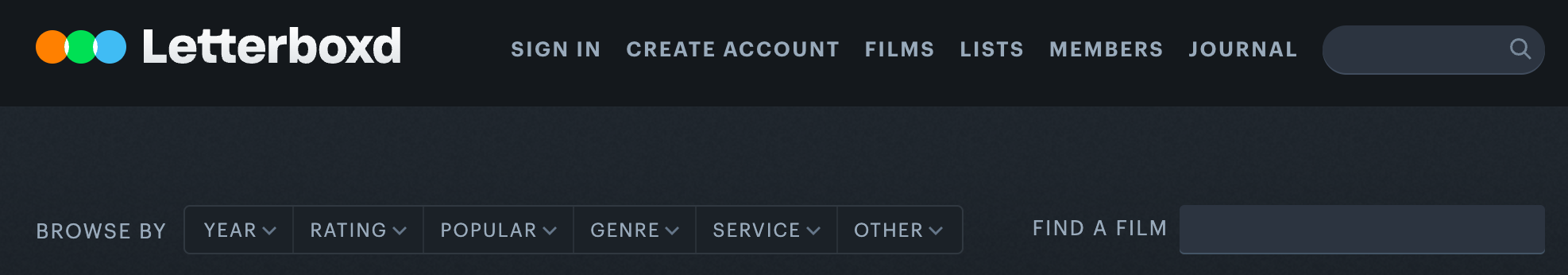}
    \caption{Screenshot of Letterboxd controls interface showing the traditional filters (genre and decade) that we compare against CTRL-Rec in our reachability experiments. We do not allow users to order by rating or popularity, as we automatically order movies using the engagement-based recommender system.}
    \label{fig:lb-controls}
\end{figure}

\subsection{Technical Details of Filter Selection}\label{app:filter-greedy}

\textbf{Genre Selection.} We analyze the target feed's genre distribution by tallying how often each genre appears across the top-10 movies, then sort genres by frequency. We construct filter combinations by incrementally adding the most common genres (e.g., Action → Action+Adventure → Action+Adventure+War), using conjunctive (AND) filters where movies must satisfy all specified genres.

\textbf{Decade Integration.} We identify the three most frequent decades in the target feed and test each genre combination both alone and with each decade individually. For example, "Action+Adventure" would be tested as: (1) Action+Adventure, (2) Action+Adventure+1990s, (3) Action+Adventure+2000s, etc. We avoid combining decades since "1990s AND 2000s" would return no results.

\subsection{Training Data Categories}\label{app:training-categories}

Our training data spans 10 categories of user preference requests designed to capture diverse ways users express movie preferences:

\begin{enumerate}
    \item \textbf{One-time:} Immediate, situational requests for specific moments (e.g., "need something short", "want to cry")
    \item \textbf{Long-term:} Persistent preferences and standing rules (e.g., core values, content boundaries)
    \item \textbf{Aspirational:} Requests for personal growth and cultural enrichment (e.g., "help me develop appreciation for...")
    \item \textbf{Changing:} Evolving tastes transitioning from old to new preferences (e.g., "I used to like X, now I want Y")
    \item \textbf{Ambiguous:} Poetic, impressionistic requests using metaphor and mood (e.g., "something gentle")
    \item \textbf{Similarity-based:} Comparative requests referencing specific films (e.g., "like Blade Runner but comedic")
    \item \textbf{Smart-filtering:} Highly specific requests requiring deep content knowledge (e.g., cinematography techniques)
    \item \textbf{Smart-filtering-easy:} Clear, specific requests in simple language (e.g., plot elements, character types)
    \item \textbf{Logical-filtering:} Precise logical filtering with boolean operators (e.g., "Horror from 1990s BUT NOT zombies")
    \item \textbf{Refinement:} Adjusting recommendations relative to recent patterns (e.g., "more diverse than lately")
\end{enumerate}

\section{User study details} \label{app:human-study}

\begin{figure}[h]
    \centering
    \includegraphics[width=1\linewidth]{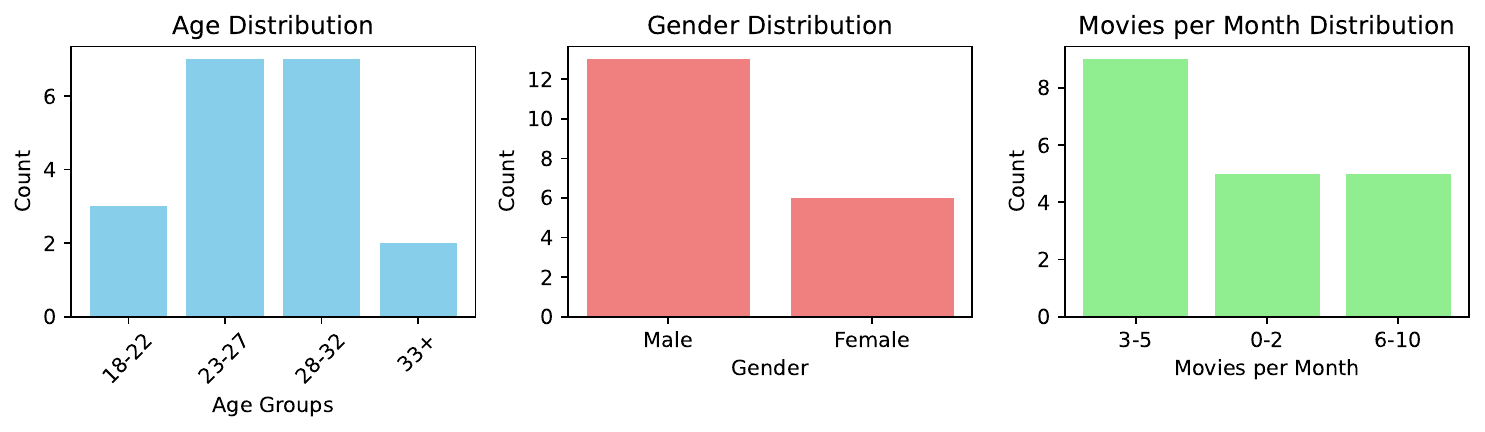}
    \caption{Distribution of age, gender, and average number of movies watched per month among Letterboxd users in our human study.} \label{fig:user_demographics}
\end{figure}

\textbf{Participants.} We conducted a human study with 19 Letterboxd users to evaluate the effectiveness of CTRL-Rec in a movie recommendation task leveraging users' own Letterboxd data. We recruited participants through social media posts, university Slack channels, and random sampling of personal connections that met the study criteria (18+ years old, have 10+ movie ratings on Letterboxd).  Participant demographics are shown in Fig. \ref{fig:user_demographics}. 6 participants identified as female and 13 as male. The mean participant age was $27.0\pm4.3$ (min: 21, max: 36). Most participants watched 3--5 movies per month on average. Our study was approved by our institution's IRB and informed consent was obtained from all participants. 

\begin{figure}
    \centering
    \includegraphics[width=0.9\linewidth]{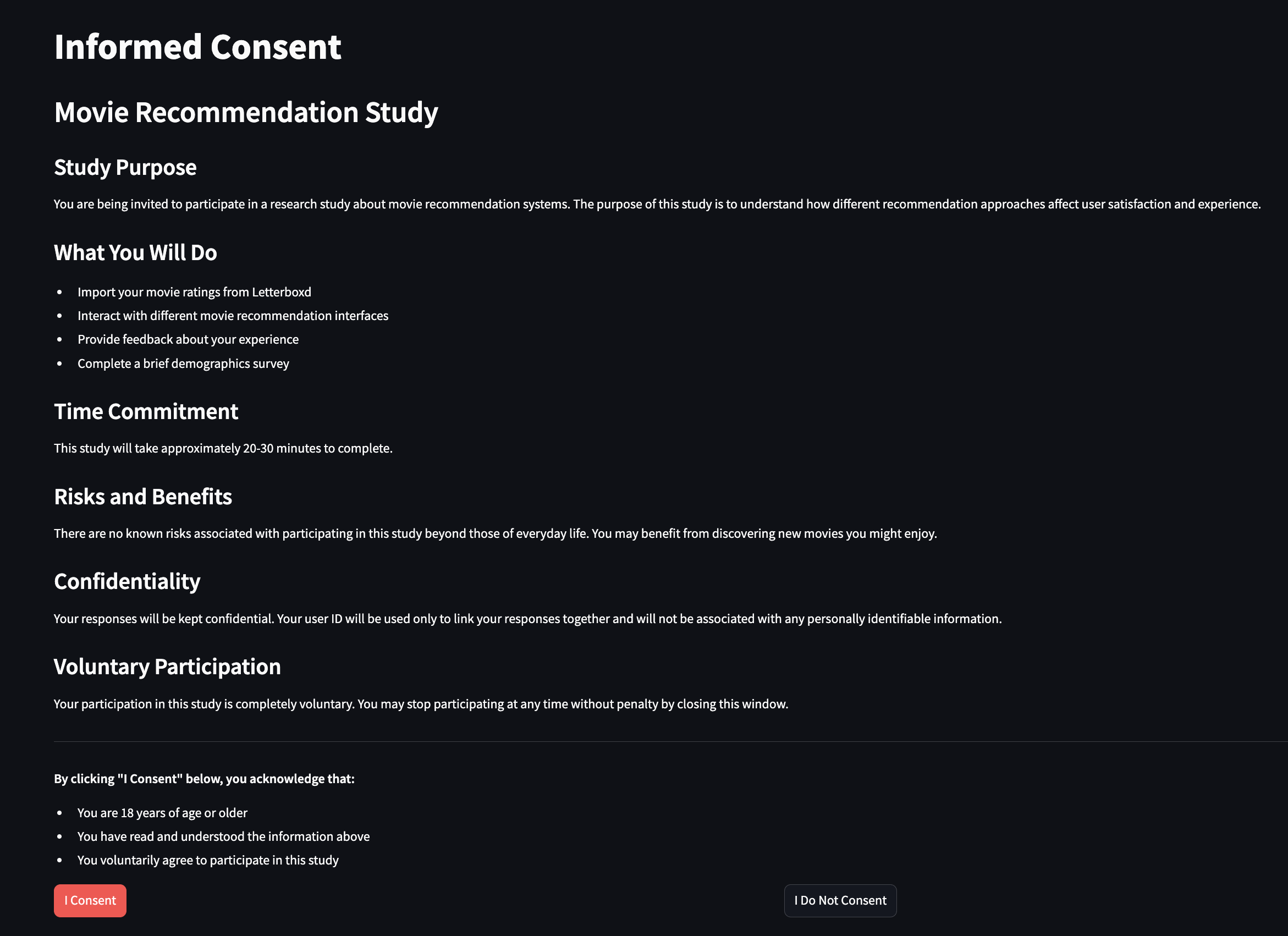}
    \caption{\textbf{User study, consent form.}} \label{fig:consent-form}
\end{figure}

\textbf{Setup.} Our study had a within-subjects design with two cunterbalanced conditions: (a) CTRL-Rec\footnote{We used a $w_\text{control}= 0.995$ for the trade-off between engagement and stated preferences (\Cref{eq:realm_score}).} + genre and decade filters, (b) genre and decade filters only. We generated an initial set of recommendations using engagement-based recommender feed based on their ratings using the recommender in \Cref{sec:reachability}. Participants were then given 8 minutes per condition to use the interface to find ideal movie recommendations. For each recommended movie, we provided two buttons participants could use to indicate their engagement to the movie: ``\textit{Interested}'' if they were interested in watching the movie or ``\textit{Watched}'' if they had already seen it. After each condition, participants answered Likert scale survey questions (on a 7-point scale) to rate their experience with the system. After both conditions, participants participated in a brief, semi-structured exit interview about their experiences across both conditions. Figures \ref{fig:consent-form}-\ref{fig:ctrl-rec-user-study} show screenshots of our custom user study interface.

\textbf{$p$-values and $q$-values.} We performed Wilcoxon signed-rank tests to compute raw $p$-values for each measure. To account for multiple comparisons, we applied the Benjamini-Krieger-Yekutieli (BKY) two-stage procedure~\citep{benjamini2006adaptive}, following the approach described by \citet{anderson2008multiple}. The BKY method provides more precise control of the false discovery rate (FDR) compared to more conservative methods such as Bonferroni correction. For each measure, we report the corresponding $q$-value, which indicates the minimum FDR at which the result is considered significant at the $\alpha = 0.05$ level. \Cref{t:p-corrected} presents the $p$-values and $q$-values for all measures. All measures with $p < 0.05$ are also significant at an FDR threshold of $q < 0.05$ (specifically, $q \leq 0.039$). Significant results are highlighted in green.

\begin{table}[h]
\centering
\begin{tabular}{p{7cm} p{2cm} p{2cm}}
\toprule 
Measure tested & $p$-value & $q$-value\\
\midrule  
\rowcolor{green!20} Satisfaction in recommendations & 0.046 & 0.039 \\
\rowcolor{green!20} Controllability of recommender & 0.046 & 0.039 \\
\rowcolor{green!20} Expressivity of recommender & 0.0002 & 0.001 \\
Effort required for quality recommendations & 0.634 & 0.268 \\
\rowcolor{green!20} Value of filters & 0.002 & 0.006 \\
Ratio of ``liked'' movies to total recommended & 0.825 & 0.308 \\
\rowcolor{green!20} Ratio of ``watched'' movies to total recommended & 0.016 & 0.027 \\
\bottomrule
\end{tabular}
\caption{Raw $p$-values and $q$-values for all measures. All results that are significant with $p < 0.05$ are also significant at a FDR threshold of $q = 0.039$ (highlighted in green).}
\label{t:p-corrected}
\end{table}

\textbf{Survey questions.} We asked users six Likert survey questions, where participants could pick from 1 (Strongly Disagree) to 7 (Strongly Agree). The following five were asked after each condition:
\begin{enumerate}
    \item \textbf{Controllability}. “It was easy to control my recommendations''
    \item \textbf{Expressivity}. “This system allowed me to articulate preferences in an expressive way.''
    \item \textbf{Effort}. “I had to put in a lot of effort to find recommendations I like.''
    \item \textbf{Satisfaction}.  “I am satisfied with my recommendations.''
    \item \textbf{Value of filters}. “I found the genre and/or decade filter valuable.''
\end{enumerate}

Finally, at the end of both conditions, one with the natural language controls and one without, we asked participants how valuable they thought having the natural language controls were.

\begin{enumerate}[start=6]
    \item \textbf{Value of natural language}. “I found the natural language controls to be valuable.''
\end{enumerate}

\begin{figure}
    \centering
    \includegraphics[width=0.9\linewidth]{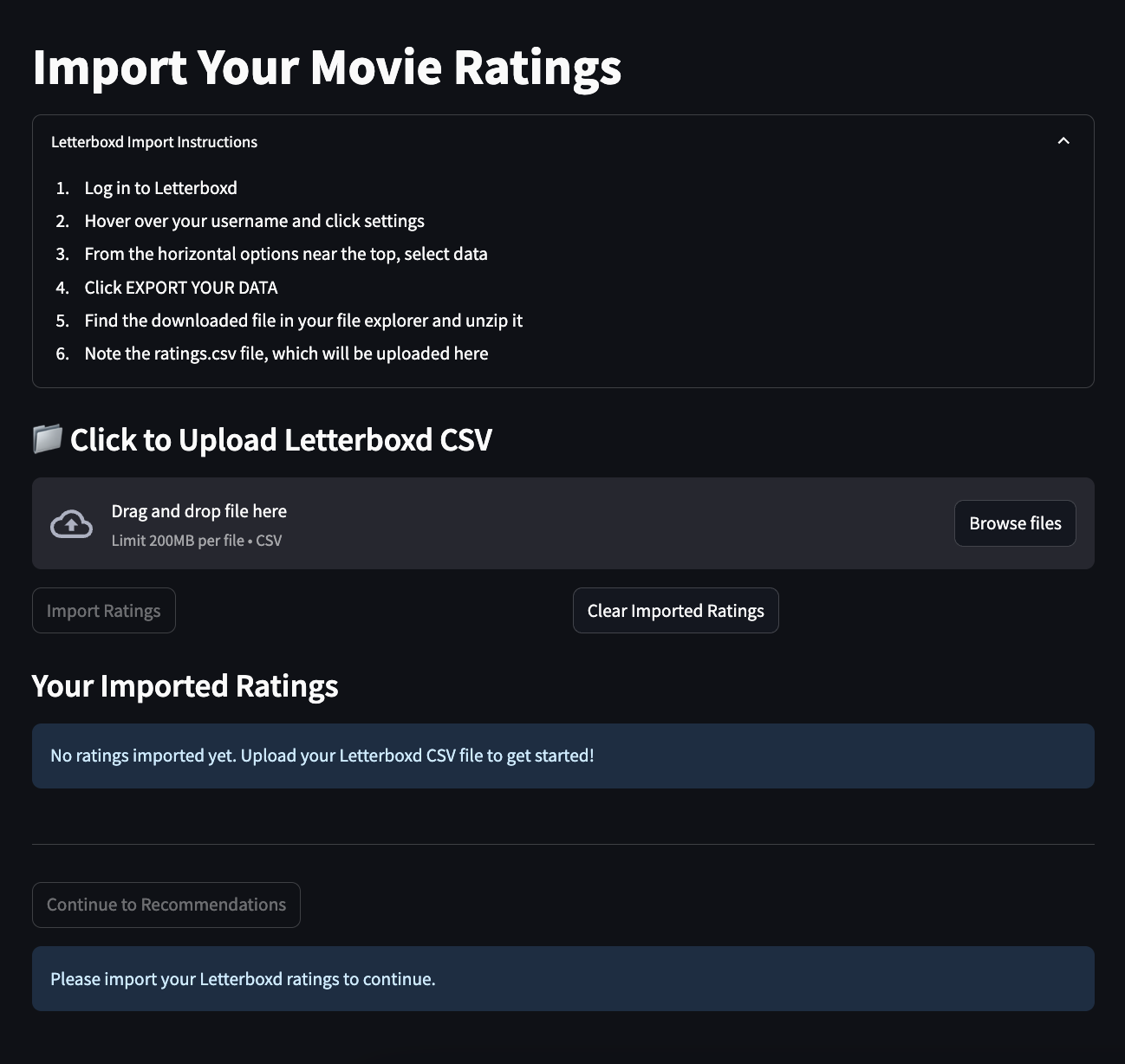}
    \caption{\textbf{User study, movie import page.}} \label{fig:movie-import-page}
\end{figure}

\begin{figure}
    \centering
    \includegraphics[width=0.9\linewidth]{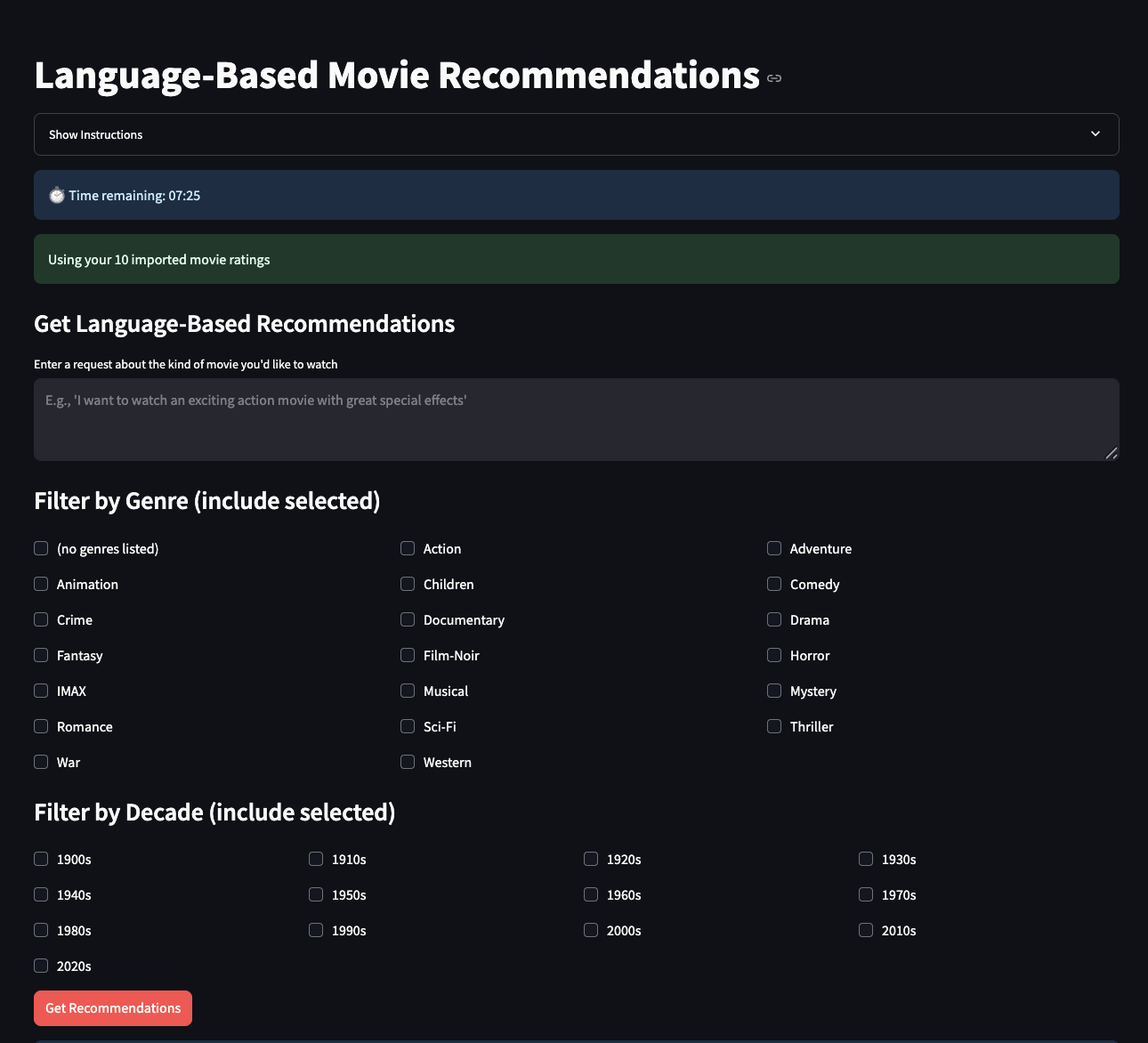}
    \caption{\textbf{User study, CTRl-Rec condition.}}
    \label{fig:ctrl-rec-user-study}
\end{figure}

\begin{figure}
    \centering    \includegraphics[width=0.9\linewidth]{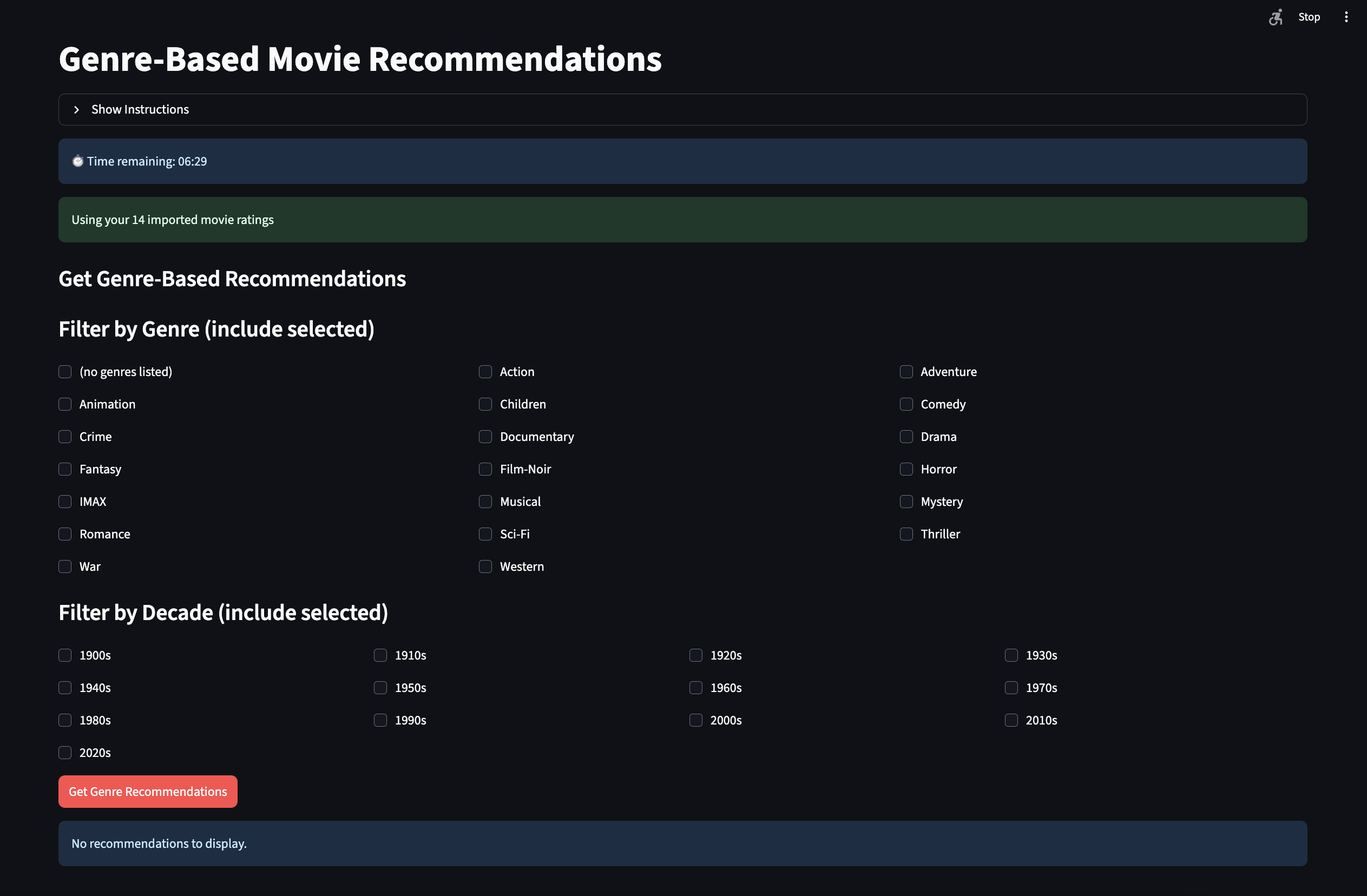}
    \caption{\textbf{User study, filters condition.}}
    \label{fig:ctrl-rec-user-study}
\end{figure}

\end{document}